\documentclass[a4paper,fleqn]{cas-sc}
\usepackage{booktabs}
\usepackage{multirow}
\usepackage{makecell}
\usepackage{amssymb}
\usepackage{amsmath}

\usepackage[table]{xcolor}
\definecolor{bestcolor}{RGB}{255, 166, 166}
\definecolor{secondcolor}{RGB}{255, 210, 166}
\definecolor{thirdcolor}{RGB}{255, 247, 187}
\newcommand{\bestcell}[1]{\cellcolor{bestcolor}#1}
\newcommand{\secondcell}[1]{\cellcolor{secondcolor}#1}
\newcommand{\thirdcell}[1]{\cellcolor{thirdcolor}#1}

\usepackage{caption}
\usepackage[authoryear]{natbib}
\let\printorcid\relax

\def\tsc#1{\csdef{#1}{\textsc{\lowercase{#1}}\xspace}}
\tsc{WGM}
\tsc{QE}

\begin{document}
\let\WriteBookmarks\relax
\def\floatpagepagefraction{1}
\def\textpagefraction{.001}

\shorttitle{SatSurfGS: Generalizable 2D Gaussian Splatting for Sparse-View Satellite Surface Reconstruction}    


\title [mode = title]{SatSurfGS: Generalizable 2D Gaussian Splatting for Sparse-View Satellite Surface Reconstruction}  

\author[1]{Min Chen} 

\author[1]{Wei Guo}
\cormark[1]

\author[1]{Bin Wang}

\author[1]{Wen Li} 

\author[2]{Tong Fang} 

\author[1]{Jinbo Zhang} 

\author[1]{Junqi Zhao} 

\author[3]{Hong Kuang}

\author[1]{Han Hu} 

\author[1]{Xuming Ge} 

\author[1]{Qing Zhu} 

\author[1]{Bo Xu}  

\address[1]{Faculty of Geosciences and Engineering, Southwest Jiaotong University, Chengdu, P.R. China}
\address[2]{School of Geography and Environment, Jiangxi Normal University}
\address[3]{College of Geodesy and Geomatics, Shandong University of Science and Technology}
\cortext[1]{Corresponding author}


\begin{abstract}
Sparse-view satellite image surface reconstruction remains highly challenging, fundamentally because the reliability of multi-view matching under satellite imaging conditions is strongly spatially heterogeneous. Affected by large photometric differences, weak textures, and repetitive textures, multi-view geometric constraints are often sparse, unevenly distributed, and locally unreliable. Although 2D Gaussian Splatting (2DGS) is more suitable than 3D Gaussian Splatting (3DGS) for the explicit representation of continuous surfaces, research on generalizable feed-forward 2DGS frameworks for sparse-view satellite surface reconstruction is still lacking. To address this issue, we propose SatSurfGS, a generalizable sparse-view surface reconstruction method for satellite imagery based on 2DGS. The proposed method builds a coarse-to-fine Gaussian attribute prediction framework and explicitly models local geometric reliability at three levels: feature learning, Gaussian parameter estimation, and training optimization. Specifically, we propose a confidence-aware monocular multi-view feature fusion module to adaptively integrate monocular priors and multi-view matching features according to local confidence; a cross-stage self-consistency residual guidance module to stabilize stage-wise Gaussian parameter refinement using the residual between the rendered height map from the previous stage and the current-stage MVS height map, together with confidence information; and a confidence bidirectional routing loss to achieve differentiated allocation of geometric and appearance supervision. Experiments on satellite datasets show that the proposed method achieves improved rendering quality, surface reconstruction accuracy, cross-dataset generalization, and inference efficiency compared with representative generalizable baselines and competitive per-scene optimization methods.

\end{abstract}





\maketitle

\section{Introduction}\label{sec1}

Three-dimensional reconstruction from satellite imagery is a key technique for acquiring fine-grained geometric information about the Earth's surface and urban buildings, and it has significant application value in urban planning, land administration, and disaster assessment. High-precision reconstruction of terrain and building surfaces can provide continuous, accurate, and structurally complete geometric representations, thereby serving as a fundamental basis for these applications \citep{r1}.

\begin{figure}
    \centering
    \includegraphics[width=0.9\linewidth]{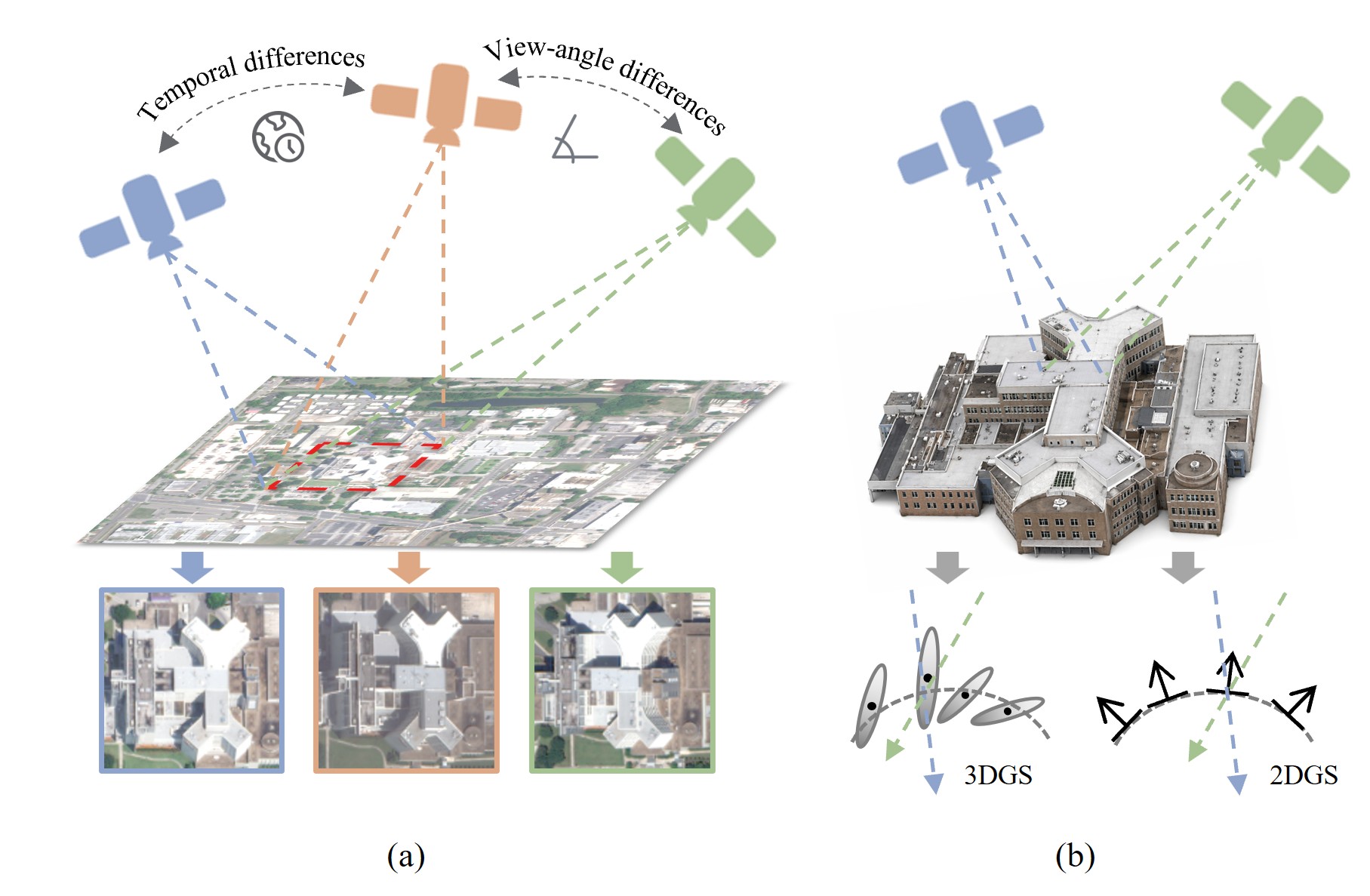}
    \caption{Imaging challenges associated with sparse satellite views and the advantages of 2DGS for surface representation. (a) Multi-view satellite images of the same area typically exhibit significant differences in phase and observation angle, resulting in marked variations in brightness and scale. (b) The surface representation of 2DGS is less prone to severe degradation than that of 3DGS, and is therefore more suitable as a basis for continuous surface reconstruction in satellite imagery.}
    \label{fig:motivation}
\end{figure}

In recent years, novel view synthesis has introduced a new paradigm for 3D scene modeling \citep{r2}. While Neural Radiance Fields (NeRF) achieves high-quality view synthesis through implicit radiance fields, Gaussian Splatting further represents scenes with explicit Gaussian primitives in an efficient and differentiable manner, achieving a better balance between rendering efficiency and detail preservation \citep{r3}. Compared with volume rendering and 3DGS, 2DGS adopts planar Gaussian primitives to better approximate continuous scene surfaces, thereby improving surface continuity, boundary sharpness, and geometric consistency \citep{r4}. For building surface reconstruction, where rooftops and facades often exhibit planar structures, clear boundaries, and locally consistent normal, the surface-oriented representation of 2DGS is highly compatible with the geometric properties of architectural scenes \citep{r5,r6}. Moreover, under satellite imaging conditions, the variation between surface normal and viewing directions is usually limited, making 2DGS less prone to severe degeneration than 3DGS, as illustrated in Fig. \ref{fig:motivation}(b). Therefore, investigating 2DGS-based surface reconstruction for satellite imagery is both necessary and promising.

However, satellite imagery is commonly affected by sparse viewpoints, large scale variations, and pronounced photometric differences, which makes robust reconstruction from limited views challenging \citep{r7}. Per-scene optimization-based Gaussian Splatting methods heavily rely on sufficient multi-view geometric constraints and long iterative fitting, but such constraints are often weakened in satellite scenarios due to limited observations and complex cross-view imaging conditions \citep{r8}. As a result, these methods may suffer from unstable convergence, low efficiency, and limited generalization. In this sense, the fundamental challenge introduced by sparse views lies not merely in the reduction of observations, but more importantly in the substantial weakening of the reliable geometric constraints on which per-scene optimization depends. In contrast, feed-forward Gaussian Splatting learns cross-scene geometric and appearance priors offline and directly predicts scene representations from sparse input images, avoiding per-scene optimization while improving reconstruction efficiency and generalization to unseen scenes\citep{r9}.

Although several generalizable feed-forward Gaussian splatting methods for sparse-view inputs have recently been proposed and have demonstrated promising reconstruction and rendering performance in indoor and UAV-based close-range scenes, most existing studies are still built upon scene assumptions of relatively rich textures, stable imaging scales, and moderate viewpoint variations. However, research on feed-forward Gaussian splatting for satellite imagery remains very limited, and there is still a clear gap in 2DGS-based continuous surface modeling. This limitation mainly stems from the following factors. Unlike close-range scenes, satellite imagery typically exhibits more pronounced radiometric differences, together with more widespread weak-texture and repetitive-texture regions. Although monocular depth estimation, empowered by large-scale pretrained models, can provide smoother and more robust structural priors, the reliability of photometric-consistency-based multi-view geometric matching remains highly spatially heterogeneous: some regions can provide stable multi-view constraints, whereas matching in other regions is prone to degradation. Consequently, there is no globally optimal fusion ratio between monocular prior features and multi-view geometric features. If these features are fused only through simple concatenation or fixed strategies, it is often difficult to adaptively allocate information sources according to local geometric reliability, and unreliable multi-view cues may even be introduced into the Gaussian parameter estimation process. Meanwhile, in an end-to-end feed-forward framework, Gaussian parameter estimation is tightly coupled with geometric priors. Once local deviations arise in the prior geometry, such errors may propagate through feature transmission or parameter initialization and subsequently interfere with the regression of current Gaussian parameters. Owing to the lack of an explicit characterization of the inconsistency between prior geometry and Gaussian parameters, the network often struggles to achieve stable and reliable parameter updates.

\begin{figure*}
    \centering
    \includegraphics[width=0.8\linewidth]{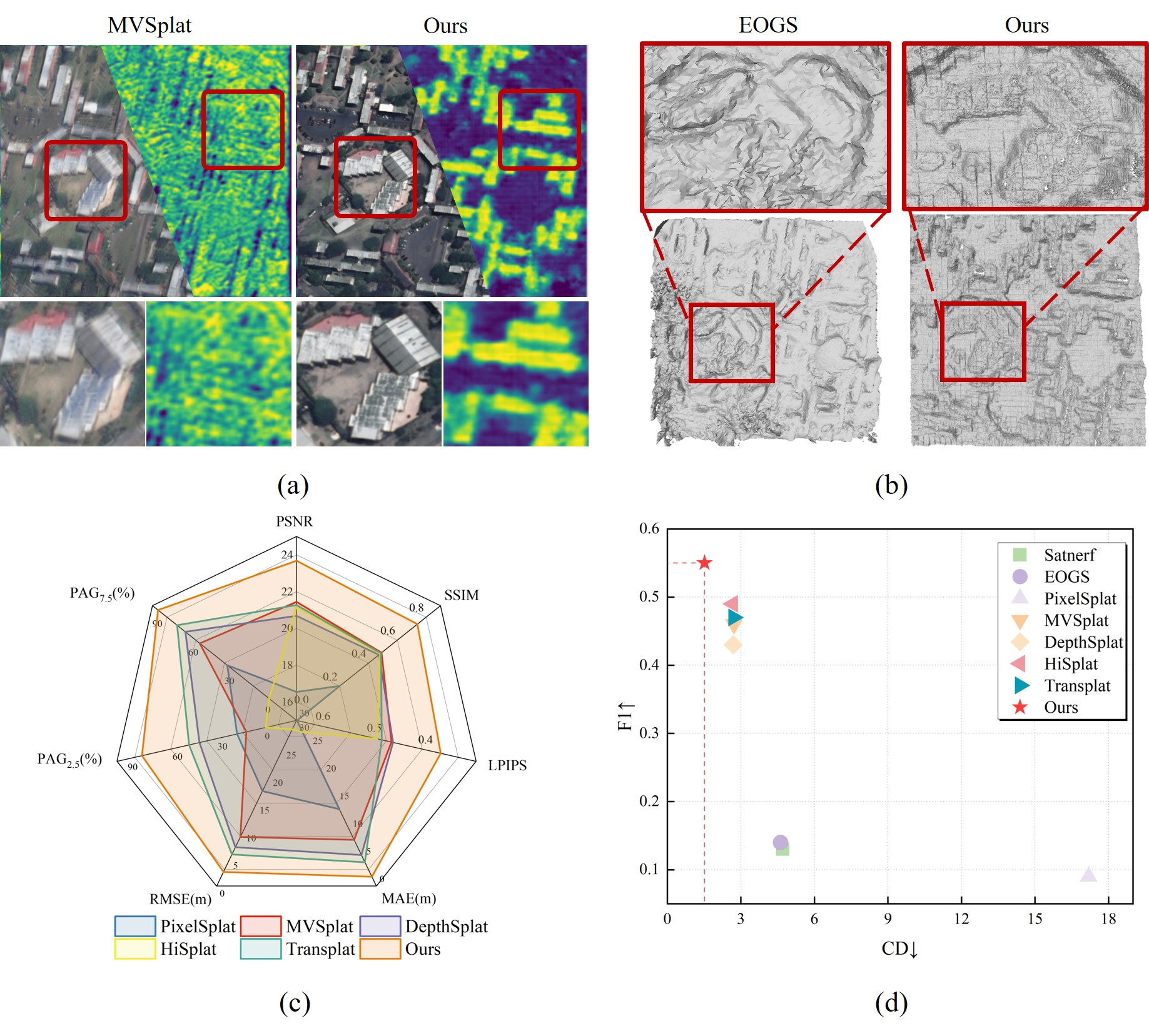}
    \caption{Comparison with existing methods: PixelSplat \citep{r9}, MVSplat \citep{r24}, DepthSplat \citep{r26}, HiSplat \citep{r27}, Transplat \citep{r25}, EOGS \citep{r8} and Sat-nerf \citep{r7}. (a) Comparison of rendered images and estimated height maps;(b) Comparison of surface reconstruction results; (c) Generalization results on the DFC19 \citep{r44}; (d) Comparison of surface reconstruction accuracy on DFC19.}
    \label{fig:Comparison}
\end{figure*}

To address the above challenges, we propose SatSurfGS, a generalizable sparse-view surface reconstruction framework for satellite imagery based on 2DGS. Under rational polynomial coefficient (RPC) geometric constraints, it performs feed-forward generalizable learning for surface reconstruction and explicitly models and handles spatially heterogeneous observation reliability from three aspects: information selection, state updating, and optimization allocation. Specifically, (1) At the feature learning stage, to address the spatial inconsistency of geometric cue reliability under sparse satellite observations, we introduce cross-scene geometric priors through a monocular depth estimator and design Confidence-Aware Monocular–Multiview Feature Fusion (CMMF). This module allows the network to adaptively select more reliable information sources at each location, thereby improving feature representation in degraded regions. (2) At the Gaussian parameter estimation stage, to reduce the interference caused by inconsistencies between existing geometric priors and current observations, we propose Cross-Stage Self-Consistency Residual Guidance (CSRG). This module explicitly models the normalized residual between the GS-rendered height map from the previous stage and the MVS-estimated height map from the current stage. It further incorporates the confidence information of the current stage to selectively filter and propagate cross-stage geometric errors, thereby supporting more stable and progressive refinement of the current surface state. (3) At the optimization stage, to address the fact that regions with different levels of geometric accuracy require different optimization priorities, we further propose Confidence Bidirectional Routing Loss (CBRL). This loss allocates supervision signals according to local confidence. It strengthens geometric correction in regions with low geometric accuracy, while placing greater emphasis on appearance reconstruction in regions where the geometry is already reliable. In this way, it reduces the mutual interference between geometric learning and appearance fitting. 

Through these designs, SatSurfGS establishes a unified reliability-driven framework for satellite surface modeling across feature representation, Gaussian parameter estimation, and overall optimization. As a result, it enables 2DGS-based surface reconstruction for sparse-view satellite imagery. An overview comparison is shown in Fig. \ref{fig:Comparison}. Compared with existing methods, the proposed SatSurfGS yields more accurate and visually coherent reconstruction results, while also achieving superior overall quantitative performance.

Our main contributions can be summarized as follows:
\begin{enumerate}
\item We propose SatSurfGS, a feed-forward 2DGS framework for sparse-view surface reconstruction from satellite imagery. This framework use the RPC camera model to estimate surface height maps and 2DGS parameters through multi-view stereo, thereby enabling end-to-end modeling from sparse satellite views to a scene surface reconstruction.
\item We introduce a unified reliability-aware mechanism for sparse-view satellite surface reconstruction. This mechanism explicitly models spatially heterogeneous geometric reliability and provides a principled way to guide information selection, stage-wise geometric refinement, and optimization allocation in a cascaded feed-forward 2DGS framework.
\item We propose three novel reliability-aware modules, namely Confidence-Aware Monocular–Multiview Feature Fusion (CMMF), Cross-Stage Self-Consistency Residual Guidance (CSRG), and Confidence Bidirectional Routing Loss (CBRL). These modules respectively improve adaptive feature fusion, cross-stage Gaussian parameter refinement, and differentiated supervision between height and appearance estimation.
\end{enumerate}

\section{Related work}
\subsection{Novel view synthesis}
Novel view synthesis (NVS) aims to render a target view at a given camera pose from a set of source images and their corresponding camera poses. As a representative early method, NeRF \citep{r2} represents a scene as a continuous radiance field parameterized by a multilayer perceptron and achieves high-quality novel view synthesis through volume rendering. This paradigm has substantially advanced 3D vision and inspired a large body of subsequent research \citep{r10,r11,r12,r13}. However, NeRF-based methods rely on dense sampling and volumetric integration, resulting in relatively low training and rendering efficiency. To improve efficiency, 3DGS \citep{r3} adopts explicit 3D Gaussian primitives for scene representation and combines them with visibility-aware rasterization, thereby substantially accelerating training and rendering while maintaining high visual quality and enabling efficient real-time NVS \citep{r14,r15}. Owing to its explicit representation, differentiable rendering, and strong appearance modeling capability, 3DGS has rapidly become an important technical route for radiance field reconstruction \citep{r16,r17,r18}. Furthermore, 2DGS \citep{r4} points out that the volumetric Gaussians used in 3DGS are not well suited for accurately modeling continuous real-world surfaces and may lead to multi-view surface inconsistency. To address this issue, 2DGS reformulates the scene representation as oriented 2D Gaussian disks and further improves surface reconstruction accuracy by incorporating perspective-corrected 2D splatting, depth distortion regularization, and normal consistency constraints \citep{r20}. Compared with 3DGS, 2DGS offers clear advantages in thin-structure recovery, geometric boundary preservation, and surface extraction, making it more suitable for continuous surface modeling tasks \citep{r21,r22,r23}. Nevertheless, existing NeRF, 3DGS, and 2DGS methods still largely rely on per-scene optimization, suffering from long training times, limited inference efficiency, and insufficient cross-scene generalization, which has in turn driven the development of generalizable feed-forward Gaussian splatting methods.

\subsection{Generalizable Gaussian Splatting}
To eliminate the high cost of per-scene optimization, recent generalizable Gaussian splatting methods have begun to directly predict Gaussian representations from sparse inputs using feed-forward networks. pixelSplat \citep{r9} was among the first to generate 3D Gaussians from image pairs in a feed-forward manner, achieving generalizable reconstruction by predicting 3D probability distributions and sampling Gaussian centers. MVSplat \citep{r24} further introduced plane-sweeping cost volumes to more stably localize Gaussian centers using multi-view geometric cues while predicting the remaining attributes. TranSplat \citep{r25} incorporates depth confidence and monocular depth priors within a Transformer framework to enhance matching in non-overlapping regions and regions with similar textures. DepthSplat \citep{r26} explicitly bridges depth estimation and Gaussian splatting, leveraging pretrained monocular depth features to improve feed-forward reconstruction quality. In addition, HiSplat \citep{r27} adopts a hierarchical coarse-to-fine Gaussian representation to balance large-scale structural modeling and fine-detail recovery. Splatter-360 \citep{r28} constructs spherical cost volumes for wide-baseline panoramic images and combines them with cross-view attention. FreeSplatter \citep{r29} further extends this paradigm to sparse-view scenarios without known poses. Overall, these methods have significantly improved reconstruction efficiency and cross-scene generalization under sparse-view settings. However, most of them are still designed for indoor, object-centric, or close-range scenes, and pay insufficient attention to the more complex imaging models, wide-baseline observations, and geometric degradation issues encountered in satellite imagery.

\subsection{Novel view synthesis for Satellite Images}
Compared with close-range visual scenes, NVS for satellite imagery faces more complex imaging conditions and geometric constraints. Satellites typically employ a push-broom imaging mechanism, which is more appropriately modeled by the RPC model. Moreover, multi-temporal satellite imagery commonly suffers from illumination variation, shadow differences, transient object interference, and sparse viewpoints, making it difficult to directly transfer classical NVS methods to remote sensing scenarios. To address these challenges, Sat-NeRF \citep{r7} was the first to introduce the RPC camera model into a satellite NeRF framework and, by incorporating shadow modeling and uncertainty-weighted strategies, achieved NVS and 3D reconstruction for multi-view satellite imagery. Subsequently, EO-NeRF \citep{r30} further improved geometric consistency and photometric modeling for multi-temporal Earth observation imagery, yielding better performance in both NVS and digital surface model (DSM) generation. SparseSat-NeRF \citep{r31} alleviated the problem of insufficient geometric constraints under sparse-view settings by introducing dense depth supervision derived from traditional stereo matching. SatensoRF \citep{r32} improved the training and inference efficiency for large-scale satellite scenes through tensor decomposition. rpcPRF \citep{r33} combined MPI representation with the RPC camera model to achieve generalizable NVS for single-image or sparse-input satellite imagery.

With the development of Gaussian Splatting, some recent studies have begun to introduce explicit Gaussian representations into satellite-image NVS and photogrammetric tasks. Compared with NeRF-based methods, Gaussian Splatting offers clear advantages in training and rendering efficiency, making it more suitable for large-scale satellite scene modeling. SatGS \citep{r34} achieves high-quality novel view reconstruction for satellite content by adapting the camera model, initialization strategy, and temporal variation. EOGS \citep{r8} further demonstrates that, after adaptation to remote sensing scenarios, the standard 3DGS framework can significantly reduce training cost while maintaining reconstruction accuracy. More recently, SkySplat \citep{r35} explicitly integrates the RPC model into a generalizable 3DGS pipeline and proposes a self-supervised reconstruction framework for multi-temporal sparse satellite imagery, with a particular focus on addressing limited geometric constraints, transient object interference, and photometric inconsistency, representing an important advance in generalizable Gaussian splatting for satellite imagery. In a further step, Skyfall-GS \citep{r36} combines 3DGS with diffusion priors to generate immersive 3D urban scenes that support real-time exploration, thereby further extending the application boundary of satellite Gaussian splatting in large-scale scene visualization. However, existing studies on satellite Gaussian Splatting still mainly focus on 3DGS and NVS, with greater emphasis on visualization quality and immersive scene generation. Systematic research on a generalizable feed-forward 2DGS framework for sparse-view satellite imagery with continuous surface reconstruction as the primary objective remains lacking. 

To this end, this paper proposes SatSurfGS, a generalizable 2DGS-based surface reconstruction method for sparse-view satellite imagery. By introducing a confidence-aware monocular–multiview feature fusion mechanism and a cross-stage self-consistency residual guidance strategy, the proposed method improves the stability of Gaussian parameter estimation and the geometric consistency of surface reconstruction in complex satellite scenes.


\section{Methods}
\subsection{Overview}

\begin{figure*}
    \centering
    \includegraphics[width=\linewidth]{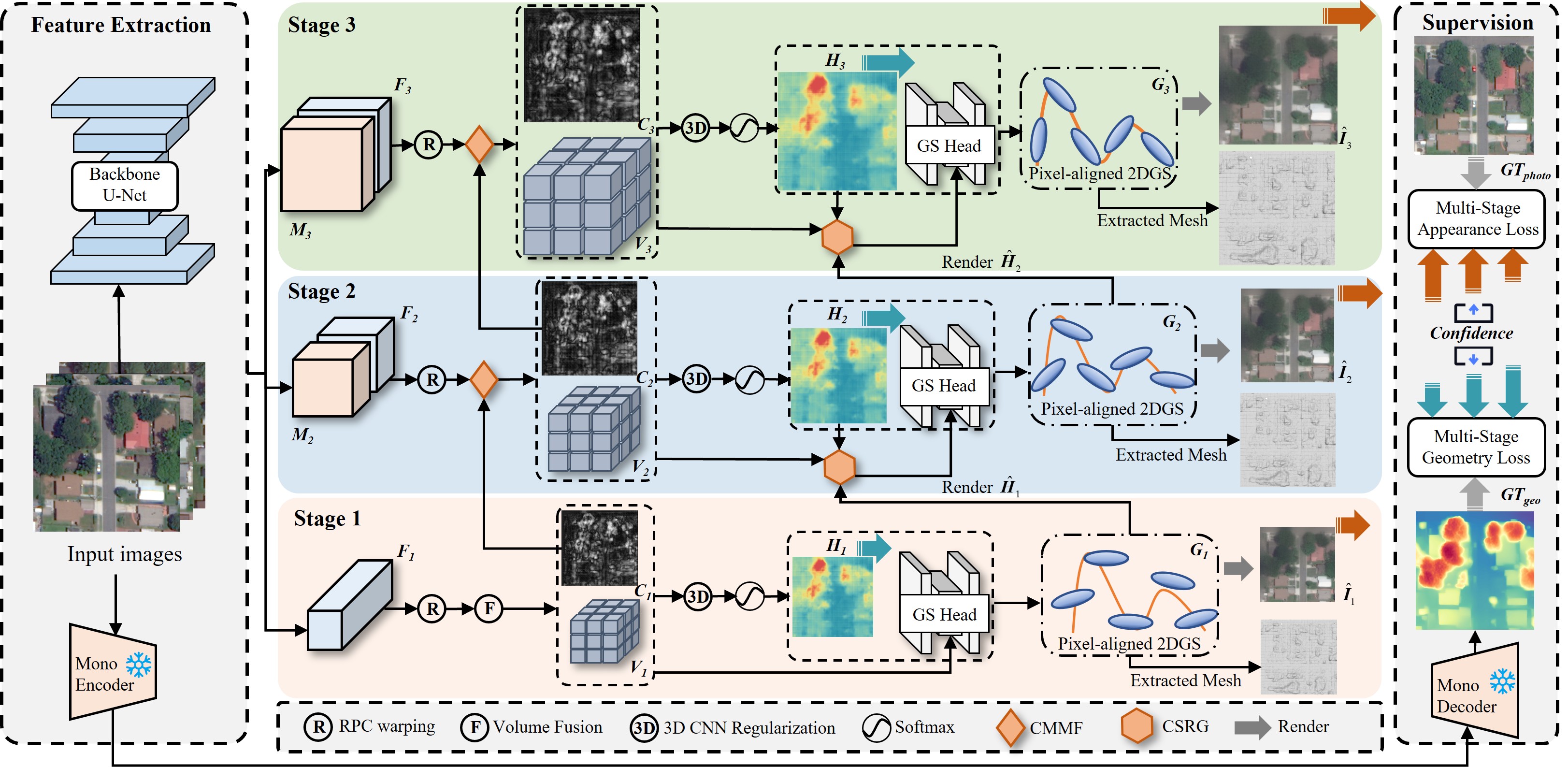}
    \caption{Overview of SatSurfGS: A Confidence-Guided Coarse-to-Fine Feed-Forward 2DGS Framework for Sparse-View Satellite Surface Reconstruction. The key designs include: (1) confidence-aware monocular–multiview feature fusion, (2) cross-stage self-cconsistency residual guidance, and (3) confidence bidirectional routing loss mechanism.}
    \label{fig:overview}
\end{figure*}

Under sparse satellite observation conditions, scene surface reconstruction is strongly affected by spatially heterogeneous geometric reliability. Because the strength of multi-view geometric constraints varies across regions, it is difficult for the network to stably model the entire scene under a unified observation assumption. Such reliability can be estimated from the cost-volume distribution: a sharp and concentrated height probability indicates confident geometric matching, while a flat or ambiguous distribution suggests uncertain height estimation. Thus, the cost-volume confidence is used as a proxy for local geometric reliability, whose definition, rationale, and empirical correlation with height errors are discussed in Sec. \ref{sec3.4}. Based on this proxy, we propose SatSurfGS, a three-stage cascaded framework for satellite multi-view Gaussian surface reconstruction, as illustrated in the Fig. \ref{fig:overview}. The framework is built upon the RPC projection model and MVS geometric inference, and uses 2DGS as the representation of continuous surfaces. By explicitly modeling local geometric reliability, it improves the robustness of satellite surface reconstruction at three levels: feature representation, Gaussian parameter estimation, and training optimization.

Specifically, first, the input images $I$ are passed through a backbone feature extraction network to obtain stage-wise multi-view feature volumes $F_s$, where $\mathrm{s}\in\{1,2,3\}$. Meanwhile, monocular prior features $M_s$ are extracted using a frozen monocular encoder. At stages with $\mathrm{s>1}$ $F_s$ and $M_s$ are first geometrically aligned across views via the RPC model and then fused by the CMMF module to obtain the fused feature volume $\tilde{F}_{s}$ for the current stage. Subsequently, the network constructs the cost volume $V_s$ from $\tilde{F}_{s}$ through volume fusion and estimates the confidence map $C_s$ for the current stage. The cost volume $V_s$ is then regularized by a 3D CNN and processed with SoftMax to produce the height map $H_s$. Based on the cost volume $V_s$ and the confidence map $C_s$, the GS decoder further predicts the pixel-aligned 2DGS representation $G_s$, from which the stage-wise RGB image $\hat{I}_{s}$ and rendered height map $\hat{H}_{s}$ are obtained. For stages with $\mathrm{s>1}$, the rendered height map from the previous stage $\hat{H}_{s-1}$ is up sampled and subtracted from the current-stage MVS-estimated height map $H_s$ to obtain the cross-stage self-consistency residual, denoted as $\Delta H_{s}$. This residual, together with the confidence map $C_s$ of the current stage, is then fed into the CSRG module to explicitly characterize cross-stage geometric inconsistency and guide the selective refinement of Gaussian parameters at the current stage. Through this coarse-to-fine three-stage cascaded optimization, the network is able to preserve stable structures in high-confidence regions while actively correcting low-confidence or geometrically inconsistent regions, ultimately producing a more accurate and consistent Gaussian scene representation $G_3$. In addition, we introduce a confidence bidirectional routing loss, in which multi-stage appearance and geometric losses are computed for the rendered RGB image $\hat{I}_{s}$ and the estimated height map $H_s$ at each stage, respectively, while confidence is used to decouple geometric refinement from appearance optimization. Specifically, the ground truth for the appearance loss $GT_{photo}$ is the original image from an additional viewpoint, whereas the ground truth for the geometric loss $GT_{geo}$ is the depth map obtained by the monocular depth estimator from the image.

\subsection{Cost Volume Construction}
To simultaneously extract cross-view geometric matching cues and stable monocular structural priors, we first adopt a dual-branch feature extraction architecture to separately derive multi-view geometric features and monocular prior features from the input source images. These features are then used in subsequent stages for cost volume construction and reliability-aware information interaction. After feature extraction, we compute the correlations among features across multiple views through differentiable RPC transformation \citep{r37}. Specifically, for stage $s$ , given a target view $i$ , a pixel location $(u, v)$, and a set of discrete height hypotheses $\{h_s^k\}_{k=1}^K$, we first use the RPC back-projection operator of the target view to recover the image pixel under the height hypothesis $h_s^k$ into a 3D point:

\begin{equation}
(\mathrm{Lat}_s^{i,k},\mathrm{Lon}_s^{i,k},\mathrm{Hei}_s^{i,k})=\mathrm{RPC}_i^{-1}(u,\nu,h_s^k),
\end{equation}
Where $(\mathrm{Lat}_s^{i,k},\mathrm{Lon}_s^{i,k},\mathrm{Hei}_s^{i,k})$ denotes the 3D geodetic coordinates corresponding to the target-view pixel $(\mathrm{u_s},\mathrm{v_s})$ under the height hypothesis $h_s^k$. We then use the RPC forward projection operator of a source view $j$ to map this 3D ground point onto the source image plane:

\begin{equation}
(u_s^{j->i,k},v_s^{j->i,k})=\mathrm{RPC}_j(\mathrm{Lat}_s^{i,k},\mathrm{Lon}_s^{i,k},\mathrm{Hei}_s^{i,k}),
\end{equation}
After obtaining the reprojected pixel location $(u_s^{j->i,k},\nu_s^{j->i,k})$, the reprojected feature aligned to the target view $\tilde{F}_s^{j->i,k}$, can be sampled from the source-view feature map $F_s$ through interpolation. Once the reprojected features from all source views are obtained, we compute the variance metric over the reprojected features from all views except the target view to construct the cost volume $V_s$ at stage $s$. The cost volume $V_s$ is then regularized by 3DCNN, followed by softmax normalization along the height dimension to obtain the probability distribution $P_s$, which represents the probability of each pixel under different height hypotheses. Based on this probability distribution, the height map $H_s$ at the current stage is regressed by expectation:

\begin{equation}
    H_s=\sum_{k=1}^KP_s\cdot h_s^k.
\end{equation}

\subsection{2D Gaussian Parameters Prediction}
Next, after obtaining the height map $H_s$, confidence map $C_s$, and cost volume $V_s$ at stage $s$, the network further predicts the pixel-aligned 2DGS representation for the current stage:

\begin{equation}
    G_s=\{\mu_s,\lambda_s,q_s,c_s,\alpha_s\},
\end{equation}
where $(\mu_s,\lambda_s,q_s,c_s,\alpha_s)$ denote the center position, scale, rotation, appearance attributes, and opacity of the stage-$s$ 2D Gaussian primitive, respectively. For the Gaussian center position $\mu_{s}$, it is directly obtained by back-projecting the estimated height value $H_s$ through the corresponding RPC model. The remaining parameters are predicted by a lightweight convolutional Gaussian head, formulated as:

\begin{equation}
    \{\lambda_s,q_s,c_s,\alpha_s\}=\Phi_s^{gs}(V_s,\tilde{F}_s,I_s,C_s,1_{s>1}\Delta H_s),
\end{equation}
where the convolutional head takes the cost volume $V_s$, the fused feature volume $\tilde{F}_s$ the original RGB image $I_s$ , and the confidence map $C_s$ as inputs. In addition, for stages with $\mathrm{s>1}$, the cross-stage self-consistency residual $\Delta H_{s}$ is further introduced in the CSRG module as an additional input; the detailed process is described in section \ref{sec3.5}. Here, $1_{s>1}$ is an indicator function: when $\mathrm{s=1}$, this term is set to 0; when $\mathrm{s>1}$, it is set to $\Delta H_{s}$. The rotation quaternion $q_s$ is derived using a pinhole camera model that approximates the RPC model \citep{r40}.

\subsection{Reliability Proxy from Cost-Volume Confidence}
\label{sec3.4}
Due to weak textures, repetitive structures, radiometric inconsistencies, and limited view-angle diversity in satellite imagery, the geometric reliability of RPC-based multi-view matching is highly spatially heterogeneous. Therefore, a local reliability indicator is required to guide subsequent feature fusion, Gaussian parameter refinement, and training supervision. In this work, we use the confidence derived from the height probability distribution of the cost volume as a proxy for local geometric reliability.

Specifically, the confidence map $C_s$ is defined as the maximum probability response along the height dimension. It characterizes the reliability of the current height estimation and, to a certain extent, also reflects the reliability of photometric consistency across multi-view images \citep{r38,r39}. The maximum probability response measures the concentration of the height posterior distribution. A sharp distribution with a dominant peak indicates that the multi-view observations consistently support a specific height hypothesis, whereas a flat or multi-modal distribution indicates ambiguous matching and larger geometric uncertainty. It is defined as:

\begin{equation}
    C_s=\max_kP_s.
\end{equation}

Compared with alternative reliability cues, the maximum probability response provides a direct and interpretable indicator of local geometric reliability. Photometric residuals are easily affected by illumination changes, shadows, seasonal variations, and non-Lambertian effects, and therefore do not necessarily indicate geometric correctness. Image gradients or texture strength can only describe local appearance richness, but cannot determine whether cross-view correspondences are geometrically consistent. Monocular uncertainty reflects the stability of learned priors, but is not directly tied to the current RPC-guided multi-view observation. Other distribution-based measures, such as entropy and probability variance, can also reflect the ambiguity of the height probability distribution. However, entropy characterizes the global uncertainty over all height candidates and may be affected by low-probability responses or multi-modal distributions, while probability variance is sensitive to the sampling range and interval of height hypotheses. By contrast, the maximum probability response focuses on the dominant supported height hypothesis and directly reflects whether the current MVS observation has a clear geometric solution. For this reason, we use it as the confidence map in the proposed framework.

\begin{figure*}
    \centering
    \includegraphics[width=0.9\linewidth]{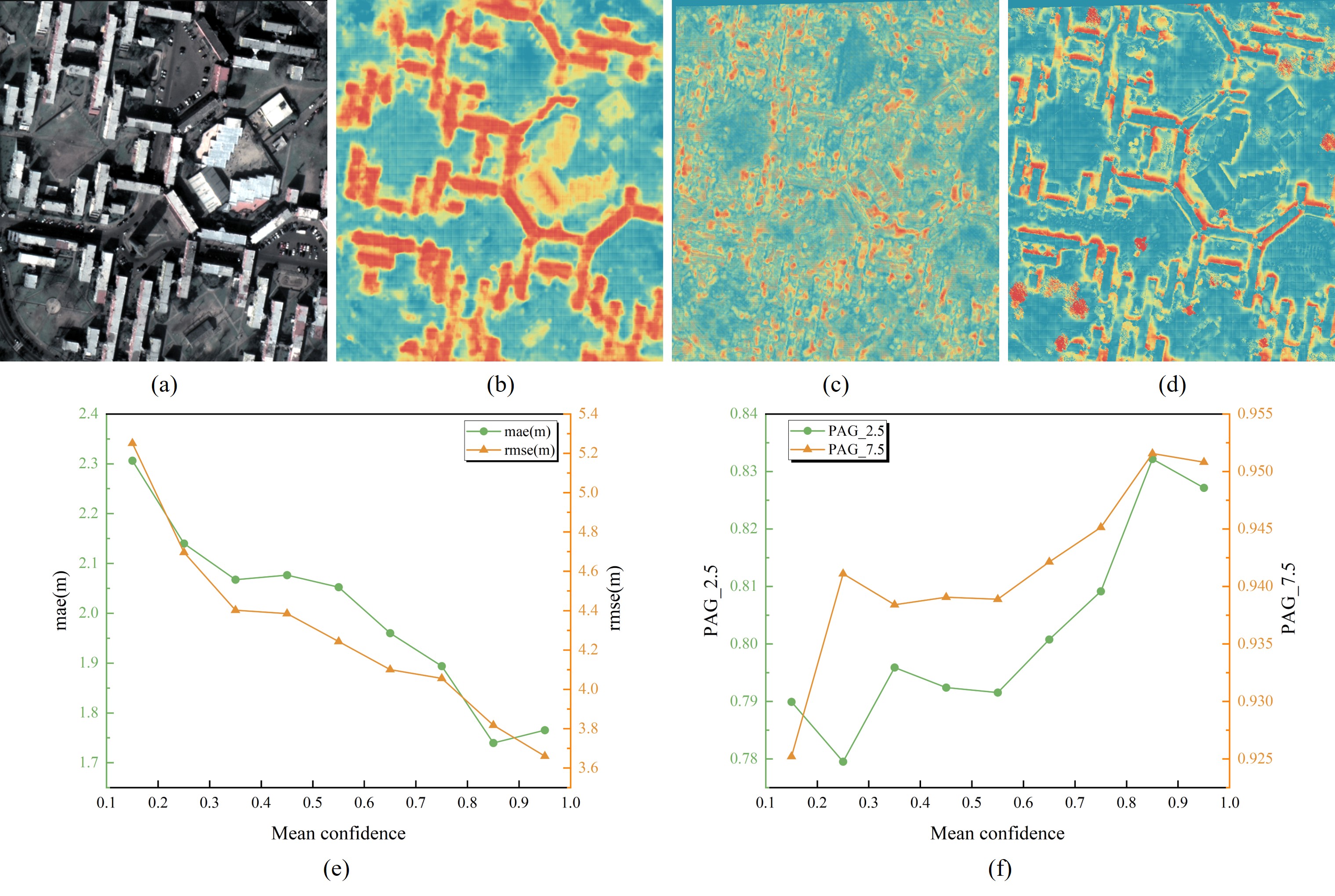}
    \caption{Visualization and confidence-binned analysis of the proposed confidence. (a) Input satellite image. (b) Estimated height map. (c) Predicted confidence map. (d) Height error map. (e) Confidence-binned MAE and RMSE. (f) Confidence-binned $PAG_{2.5}$ and $PAG_{7.5}$. The spatial comparison between (c) and (d) shows that low-confidence regions generally coincide with regions of larger geometric error. The binned statistics in (e) and (f) further generally show that higher-confidence bins are associated with lower geometric errors and better reconstruction accuracy, supporting the use of the predicted confidence as a reasonable proxy for local reconstruction reliability.}
    \label{fig:confidenceBins}
\end{figure*}

To examine whether the predicted confidence can indeed serve as a meaningful reliability indicator, we further visualize its spatial distribution and analyze its relation to geometric quality, as shown in Fig. \ref{fig:confidenceBins}. From the spatial maps in Fig. \ref{fig:confidenceBins}(c) and Fig. \ref{fig:confidenceBins}(d), it can be observed that regions with lower confidence tend to coincide with regions of larger height error, especially around structurally complex areas, local boundaries, and geometrically ambiguous regions. This indicates that the proposed confidence is not randomly distributed, but is able to reflect the local difficulty and uncertainty of geometric estimation to a certain extent. Moreover, the confidence-binned statistics in Fig. \ref{fig:confidenceBins}(e) and Fig. \ref{fig:confidenceBins}(f) further show a consistent global trend: as the mean confidence increases, both MAE and RMSE generally decrease, while $PAG_{2.5}$ and $PAG_{7.5}$ generally increase. These observations suggest that the estimated confidence is positively associated with reconstruction quality and can provide a reasonable basis for reliability-aware decision making in the network.

Based on this property, the three confidence-aware components in SatSurfGS can be interpreted under a unified reliability modeling perspective. Specifically, CMMF instantiates reliability modeling at the feature level, using confidence to regulate the relative contribution of monocular prior features and multi-view matching features. CSRG instantiates reliability modeling at the state level, where confidence, together with the cross-stage geometric residual, guides the network to preserve stable Gaussian states in reliable regions while actively correcting inconsistent states in unreliable regions. CBRL instantiates reliability modeling at the supervision level, assigning different optimization priorities to regions of different reliability. In this sense, confidence serves as a shared reliability descriptor throughout the framework, consistently governing feature fusion, stage-wise state refinement, and supervision routing.

\subsection{Confidence-Aware Monocular–Multiview Feature Fusion}
\begin{figure}
    \centering
    \includegraphics[width=0.9\linewidth]{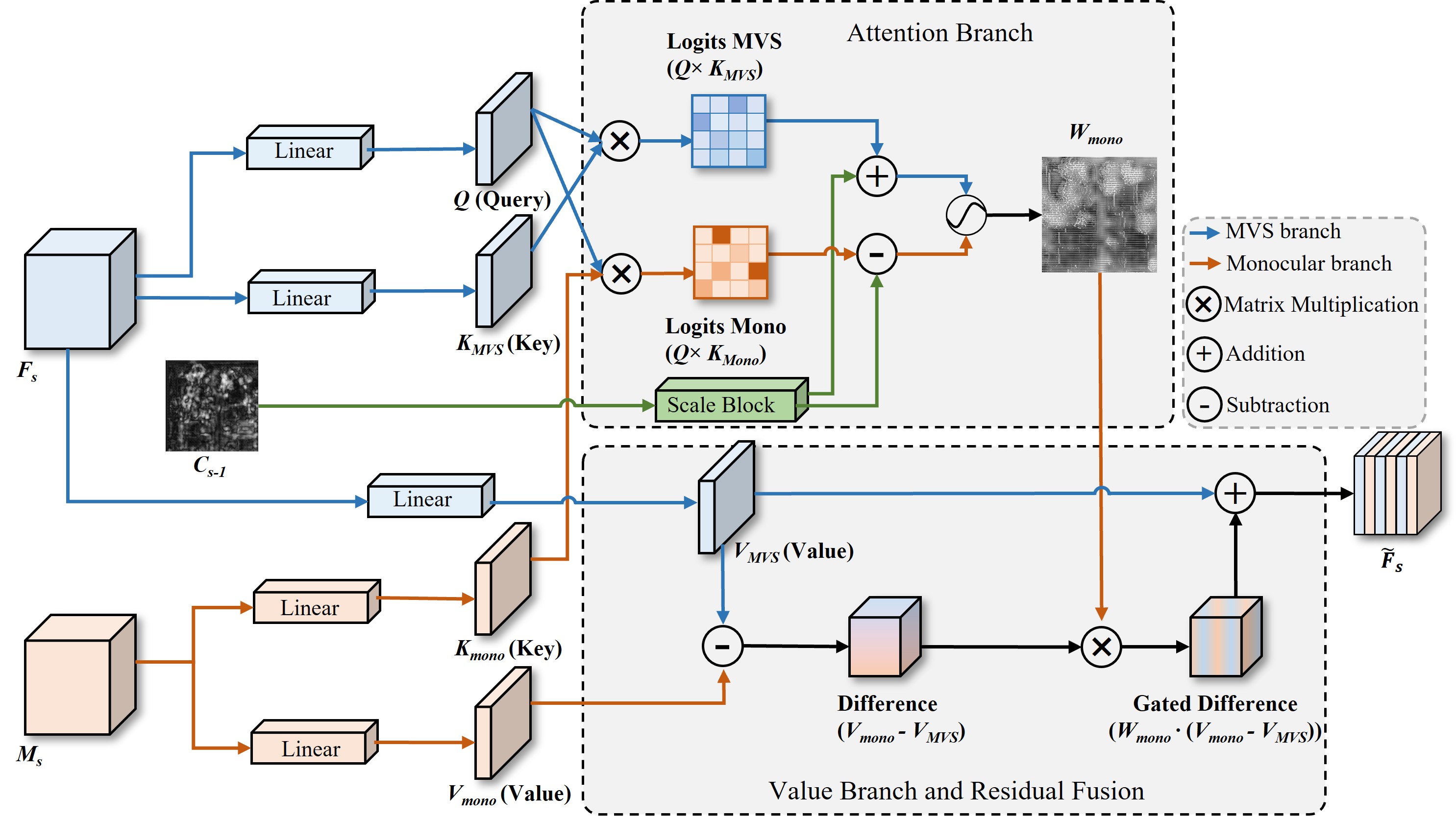}
    \caption{The illustration of confidence-aware monocular–multiview feature fusion module. The module uses confidence-guided routing to adaptively fuse multi-view and monocular features, improving robustness in ambiguous regions.}
    \label{fig:CMMF}
\end{figure}

In sparse satellite image reconstruction, the core challenge at the feature stage lies in the spatial heterogeneity of geometric constraint reliability. Multiview matching features possess strong geometric discriminability in regions with sufficient local texture and high cross-view consistency, but they tend to degrade in the presence of non-Lambertian reflection, shadow occlusion, temporal variation, and weak-texture areas. Although monocular priors can provide continuous and stable structural cues, they lack strict multi-view geometric constraints. Therefore, the key to feature fusion is not to simply combine these two types of information, but to adaptively determine the information source according to local geometric reliability. Based on this insight, we propose the CMMF module, which uses explicit confidence as guidance to dynamically allocate the contributions of multi-view features and monocular priors across spatial locations, thereby improving the robustness and accuracy of subsequent Gaussian parameter estimation. The overall structure of this module is illustrated in the Fig. \ref{fig:CMMF}.

Specifically, at stage $s$ and for source view $j$, given the feature $F_s^j$ extracted by the multi-view matching branch, the monocular feature $M_s^j$, and the confidence map $C_{s-1}$ from stage $s-1$, CMMF first uses three $1\times1$ convolutions to generate the query term and the source-specific key–value terms, respectively:

\begin{equation}
    \begin{aligned}&Q_{s}^{j}=\phi_{q}(F_{s}^{j}),\\&K_{s,mvs}^{j}=\phi_{km}(F_{s}^{j}),\\&K_{s,mono}^{j}=\phi_{ko}(M_{s}^{j}),\\&V_{s,mvs}^{j}=\phi_{vm}(F_{s}^{j}),\\&V_{s,mono}^{j}=\phi_{vo}(M_{s}^{j}),\end{aligned}
\end{equation}
Where $Q_s^j$ denotes the query term, $K_{s,MVS}^{j}$ and $V_{s,MVS}^{j}$ are the key and value terms of the MVS branch, $K_{s,mono}^{j}$ and $V_{s,mono}^{j}$ are the key and value terms of the monocular branch. Here,  $\phi_{q}$, $\phi_{km}$, $\phi_{ko}$, $\phi_{vm}$ and $\phi_{vo}$ denote learnable linear projection functions implemented by $1\times1$ convolutions, which are used to generate the query, key, and value representations, respectively. The module then computes the similarity between $Q_s^j$ and the features from the two branches, and further constructs the attention logits of the two branches by incorporating confidence-dependent bias terms:

\begin{equation}
\begin{gathered}L_{\mathrm{s,mvs}}^{j}=\frac{\left\langle Q_{s}^{j},K_{s,\mathrm{mvs}}^{j}\right\rangle}{\sqrt{d}}+\beta(\overline{C}_{s-1}),\\L_{s,mono}^{j}=\frac{\left\langle Q_{s}^{j},K_{s,mono}^{j}\right\rangle}{\sqrt{d}}-\beta(\overline{C}_{s-1}),\end{gathered}
\end{equation}
Where $d$ denotes the embedding dimension of the query/key, $\langle\cdot,\cdot\rangle$ represents the dot-product operation along the channel dimension, and $\beta(\bar{{C}}_{s-1})$ denotes the bias term generated from $\bar{{C}}_{s-1}$, which is obtained by up sampling the confidence map from the previous stage. It can be written as:

\begin{equation}
    \beta(\overline{{C}}_{s-1})=\lambda(\overline{{C}}_{s-1}-\tau),
\end{equation}
Where $\lambda$ and $\tau$ denote the scaling factor and the shifting threshold, respectively. It can thus be seen that when the confidence at a given location is high, the attention score of the multi-view branch is enhanced while that of the monocular branch is suppressed; conversely, when the confidence is low, the module tends to introduce more monocular prior information.

After obtaining the dual-branch scores, we perform SoftMax normalization along the source-branch dimension to obtain the spatial weight map $W_{s,mono}^j$ for the monocular branch. Finally, rather than directly using this weight map to compute a linear average of the two branches, CMMF adopts a residual-style routing fusion strategy:

\begin{equation}
    \tilde{F}_s^j=V_{s,\mathrm{mvs}}^j+\left(W_{s,\mathrm{mono}}^j\right)\odot\left(V_{s,mono}^j-V_{s,mvs}^j\right),
\end{equation}
Where $\odot$ denotes element-wise multiplication, and $\tilde{F}_s^j$ represents the fused feature volume. According to this formulation, $V_{s,mvs}^j$ is treated as the primary feature, while the monocular branch does not directly replace the multi-view feature; instead, it is introduced in the form of a corrective residual $(V_{s,mono}^j-V_{s,mvs}^j)$. When the monocular weight is small, the output degenerates to a representation close to pure MVS; when matching becomes unreliable in certain regions, the monocular feature can supplement the deficiencies of multi-view features in a more controlled manner. It is worth noting that the value projection layer of the monocular branch is initialized to zero, so that the module behaves approximately as an identity mapping at the beginning of training, thereby preventing the monocular prior from causing excessive disturbance to multi-view geometry in the early training stage.

Overall, the role of CMMF can be summarized in two aspects. First, by explicitly incorporating confidence, it establishes an adaptive routing mechanism that preserves MVS features in reliable regions while introducing monocular corrections in challenging regions, thereby avoiding the information contamination caused by static fusion. Second, it fuses the two types of features through residual correction rather than direct replacement, allowing monocular priors to primarily serve as complementary and corrective cues. In this way, the network enhances its robustness in weak-texture regions and areas with complex photometric variations while preserving the geometric discriminability of multi-view features. The fused features are then fed into the subsequent cost volume construction and Gaussian parameter prediction modules, providing a more stable feature foundation for the entire cascaded reconstruction process.

\subsection{Cross-Stage Self-Consistency Residual Guidance}
\label{sec3.5}
\begin{figure}
    \centering
    \includegraphics[width=\linewidth]{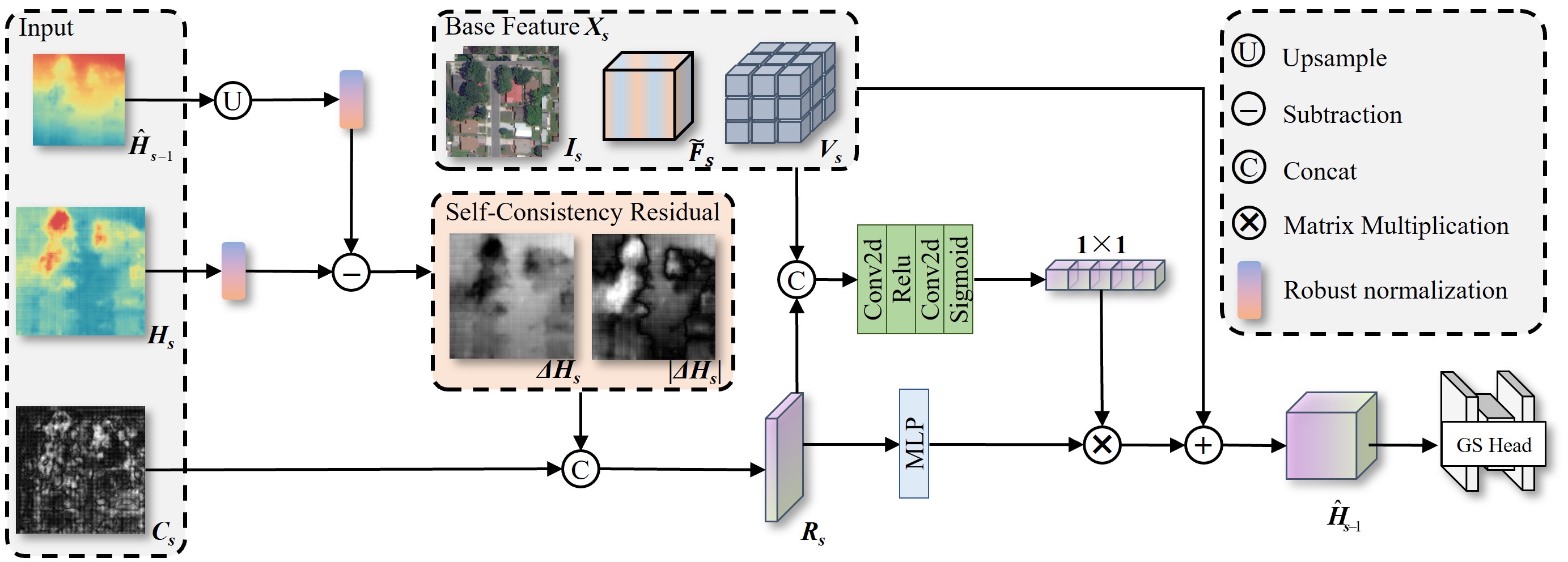}
    \caption{The illustration of cross-stage self-consistency residual guidance module. The module exploits cross-stage depth residuals and confidence to guide adaptive Gaussian refinement and improve geometric consistency across stages.}
    \label{fig:CSRG}
\end{figure}

During multi-stage Gaussian parameter estimation, the previous stage provides a coarse but globally constrained geometric state, while the current stage is responsible for further refining local surface structures. However, under sparse satellite observations, the reliability of multi-view geometry is spatially heterogeneous, and the local height estimation at the current stage is not always stable or reliable. If the subsequent prediction relies only on the current-stage observations without an explicit reference to the structural state from the previous stage, local noise and mismatches can be directly propagated into Gaussian parameter updates, leading to discontinuous geometric corrections across stages and unstable changes in Gaussian center positions and local surface geometry. Therefore, the key to cross-stage constraint is not simply to propagate previous results, but to explicitly identify cross-stage geometric inconsistency and selectively exploit such discrepancy information according to the current local reliability. To this end, we propose the CSRG module, which computes the normalized residual between the GS-rendered height map from the previous stage and the MVS-estimated height map from the current stage, and further combines it with the confidence of the current stage to construct a residual guidance signal. In this way, the Gaussian parameter estimation process at the current stage can be stabilized. The overall structure of the module is illustrated in the Fig. \ref{fig:CSRG}.

Specifically, let the predicted height map and confidence map at stage $s$ be denoted by $H_s\in\mathbb{R}^{H_s\times W_s\times1}$  and $C_s\in\mathbb{R}^{H_s\times W_s\times1}$,respectively, and let the Gaussian-rendered height map from stage $s-1$ be denoted by $\hat{H}_{s-1}\in\mathbb{R}^{H_{s-1}\times W_{s-1}\times1}$. Directly computing the difference between $\hat{H}_{s-1}$ and $H_s$ cannot stably characterize the cross-stage geometric deviation. The reason is that, although the height predictions at different stages describe the same scene, their value distributions may still be affected by the scale range, local extrema, and prediction bias. Without normalization, the residual would simultaneously contain both true geometric discrepancies and distribution-scale differences, thereby blurring its physical meaning. To address this issue, we apply robust normalization to the rendered height map from the previous stage and the current-stage height map, respectively:

\begin{equation}
    \begin{aligned}&\tilde{H}_{s-1}=\mathcal{N}(\hat{H}_{s-1}),\\&\tilde{H}_{s}=\mathcal{N}(H_{s}),\\&\mathcal{N}(X)=\frac{X-\mathrm{Median}(X)}{\mathrm{MAD}(X)+\epsilon},\end{aligned}
\end{equation}
Where $\mathrm{Median}(X)$ denotes the median, $\mathrm{MAD}(X)$ denotes the median absolute deviation, and $\epsilon$ is a small constant used to prevent division by zero. Through this treatment, the height maps from different stages are mapped into a more comparable value space, so that the subsequent residual focuses more on reflecting local geometric inconsistency rather than overall bias or scale differences. After obtaining the normalized height maps, we construct the cross-stage height residual as:

\begin{equation}
    \Delta H_s=\tilde{H}_s-\mathrm{u}(\tilde{H}_{s-1}),
\end{equation}
We further take its absolute value $\left|\Delta H_{s}\right|$, where $\mathrm{u}(\cdot)$ denotes the up sampling operation. Here, $\Delta H_{_s}$ characterizes the direction of the geometric deviation of the current stage relative to the geometric prior from the previous stage, while its absolute value reflects the magnitude of such deviation.

However, cross-stage residuals alone are still insufficient to determine whether a region should be aggressively corrected, since a large residual may originate either from a genuine geometric discrepancy or from unreliable predictions at the current stage itself. To address this issue, we further introduce the confidence map $C_s$ of the current stage as a third type of guidance signal:

\begin{equation}
    R_s=\mathrm{Concat}(\Delta H_s,\left|\Delta H_s\right|,C_s),
\end{equation}
Where $R_s\in\mathbb{R}^{H_s\times W_s\times3}$ denotes the residual-guided Residual Guidance Map. It is then projected by a lightweight embedding branch to the same channel dimension as the base feature $X_s$. Meanwhile, to make the guidance injection adaptive, we learn a positional modulation weight from the base feature of the current stage. This weight is subsequently used to perform element-wise modulation on the guidance embedding, forming an attention-controlled residual injection term. The resulting feature is finally used as the input feature $\tilde{X}_{s}$ for Gaussian estimation. This process can be formulated as follows:

\begin{equation}
    \begin{aligned}&X_{s}=\mathrm{Concat}(I_{s},\tilde{F}_{s},V_{s}),\\&\tilde{X}_{s}=X_{s}+\sigma(\phi_{w}(X_{s}))\odot(\phi_{r}(R_{s})),\end{aligned}
\end{equation}
Where $\phi_{w}(\cdot)$ denotes the weight-generation branch composed of several convolutions and nonlinear activations, $\sigma(\cdot)$ denotes the Sigmoid function, and $\phi_r(\cdot)$ is implemented as a lightweight channel embedding network.

In summary, the CSRG module aligns, normalizes, and models the residual between the rendered height map from the previous stage and the height map estimated at the current stage. It then constructs a cross-stage residual guidance map that integrates directional, magnitude, and confidence information. This guidance map is directly injected into the Gaussian estimation head of the current stage through attention-controlled residual modulation. With this design, implicit stage-wise information transfer is reformulated as explicit geometric relationship modeling. As a result, the module provides more stable and more discriminative guidance for subsequent Gaussian parameter refinement.

\subsection{Confidence Bidirectional Routing Loss}
During multi-stage cascaded optimization, prediction reliability is spatially heterogeneous. In regions where the geometry remains inaccurate, applying strong appearance supervision too early may interfere with geometric correction and may even cause geometric errors to be implicitly preserved through appearance fitting. In contrast, in regions where the geometry has already become relatively stable, repeatedly enforcing strong geometric constraints brings limited benefit. In such cases, more emphasis should be placed on further improving appearance quality. Based on this observation, we propose a confidence-based bidirectional routing loss that allocates supervision signals to the geometric branch and the appearance branch according to the prediction reliability at the current stage. This design enables targeted geometric correction and appearance refinement throughout the cascaded optimization process. The overall training loss of SatSurfGS is jointly defined by the cascaded supervision from all three stages. At each stage, the loss contains both geometric and appearance terms, and the overall optimization objective is formulated as follows:

\begin{equation}
    \mathcal{L}=\sum_{s=0}^2\omega_s\left(\lambda_{geo}\mathcal{L}_{\mathrm{geo}}^{(s)}+\lambda_{app}\mathcal{L}_{\mathrm{app}}^{(s)}\right),
\end{equation}
Where $\omega_{s}$ denotes the loss weight of each stage, $\mathcal{L}_{\mathrm{geo}}^{(s)}$ and $\mathcal{L}_\mathrm{app}^{(s)}$ represent the geometric loss and appearance loss at stage $s$, respectively. The corresponding weights are $\lambda_{geo}$ and $\lambda_{app}$.

For the geometric branch, the supervision region is jointly defined by the valid mask of the target height and the low-confidence condition. Accordingly, the geometric supervision mask is defined as:

\begin{equation}
    Mask_{geo}^{(s)}=Mask_{gt}^{(s)}\odot1(C_s<\tau),
\end{equation}
Where $\tau$ denotes the confidence threshold, $Mask_{gt}$ is the valid mask of the height ground truth, and $1(.)$ is the indicator function, which takes the value 1 when the condition is satisfied and 0 otherwise. Furthermore, we employ a multi-scale patch-level Pearson correlation loss \citep{r41} to simultaneously constrain local structural variations and the overall geometric trend. The geometric loss can thus be written as:

\begin{equation}
    \mathcal{L}_{\mathrm{geo}}^{(s)}=Mask_{\mathrm{geo}}^{(s)}\odot\frac{1}{\mid A_{s}\mid}\sum_{a\in A_{s}}\left(\lambda_{\mathrm{loc}}\mathcal{L}_{\mathrm{loc}}^{(s,a)}+\lambda_{\mathrm{glo}}\mathcal{L}_{\mathrm{glo}}^{(s,a)}\right),
\end{equation}
Where $A_s$ denotes the set of patches at stage $s$. $\mathcal{L}_{\mathrm{loc}}^{(s,a)}$ computes the local Pearson correlation within each patch to preserve the consistency of local surface undulations and depth patterns, whereas $\mathcal{L}_{\mathrm{glo}}^{(s,a)}$ introduces global statistics under the same patch partition to constrain the overall variation trend at that scale.

Correspondingly, the appearance branch is mainly responsible for improving image reconstruction quality in high-confidence regions, and the appearance supervision mask is defined as:

\begin{equation}
    Mask_{app}^{(s)}=1(C_s>\tau)
\end{equation}

On this basis, the appearance loss at stage $s$ consists of a color term, a structural term, and a perceptual term:

\begin{equation}
    \mathcal{L}_{\mathrm{app}}^{(s)}=Mask_{app}^{(s)}\odot(\lambda_{\mathrm{rgb}}\mathcal{L}_{\mathrm{rgb}}^{(s)}+\lambda_{\mathrm{ssim}}\mathcal{L}_{\mathrm{ssim}}^{(s)}+\lambda_{\mathrm{lpips}}\mathcal{L}_{\mathrm{lpips}}^{(s)}),
\end{equation}
Where $\mathcal{L}_{\mathrm{rgb}}^{(s)}$, $\mathcal{L}_{\mathrm{ssim}}^{(s)}$ and $\mathcal{L}_{\mathrm{lpips}}^{(s)}$ denote the mean squared error loss, structural similarity (SSIM) loss \citep{r42} and learned perceptual image patch similarity (LPIPS) loss \citep{r43} , respectively. $\lambda_{\mathrm{rgb}}$, $\lambda_{\mathrm{ssim}}$ and $\lambda_{\mathrm{lpips}}$ are their corresponding weights.

Overall, the proposed loss design establishes an explicit region-wise allocation mechanism through confidence: the geometry loss performs geometric correction in low-confidence regions, whereas the appearance loss focuses on appearance reconstruction in high-confidence regions. The two losses work collaboratively across the three-stage cascade, ultimately achieving simultaneous improvements in geometric accuracy and rendering quality.

\section{Experiments}
\subsection{Experimental Setup}
\subsubsection{Datasets}

The experiments were conducted on the DFC19 \citep{r44} and MVS3D \citep{r45} datasets, which contain challenging satellite multi-view scenes with complex urban structures and significant height variations. We trained and evaluated our model on the DFC19 dataset, which consists of multi-temporal imagery from Jacksonville (JAX) and Omaha (OMA), with an image size of $2048\times2048$ and a GSD of 0.3m. Following the dataset preprocessing procedure of MVSPlat and SkySplat, we selected 81 scenes for training and 15 scenes for validation. For each scene, three views were chosen as the source images, and each image was cropped into 64 patches of size 256×256. For the Cross-Dataset Generalization evaluation, consistent with SkySplat, we used three AOIs from the MVS3D dataset for validation, namely IAPRA\_001, IAPRA\_002, and IAPRA\_003.

\subsubsection{Implementation Details}
The training process was conducted on a server equipped with six NVIDIA$\mathrm{®}$ GeForce RTX 4090 GPUs (24GB VRAM each), using Python 3.10 and PyTorch 2.1.2+cu118. All models were trained for 30 epochs using the AdamW optimizer, with a batch size of 4 and a learning rate of $1\times10^{-3}$ . In addition, based on the DSM ground truth provided by the DFC19 dataset, we computed the terrain height range $[h_{min}, h_{max}]$ for each scene. During training, the scaling factor and shifting threshold in the confidence bias term were set to 1 and 0.2, respectively. The stage weights $\omega_{s}$ in the loss function were set to 0.5, 1.0, and 2.0 for the three stages, respectively.$\lambda_{geo}$ and $\lambda_{app}$ were set to 0.01 and 1.0, while $\lambda_{rgb}$, $\lambda_{ssim}$ and $\lambda_{lpips}$ were set to 1.0, 0.05, and 0.1, respectively.

\subsubsection{Evaluation Metrics}
All evaluation metrics used in this study can be grouped into three categories: rendering metrics, height-map accuracy metrics, and surface reconstruction accuracy metrics. The rendering metrics include: Peak Signal-to-Noise Ratio (PSNR) \citep{r42}, SSIM and LPIPS. The height-map accuracy metrics include: absolute error (MAE), root mean square error (RMSE), and percentage of accurate grids in total (PAG), For example, $PAG_{2.5}$ represents the ratio of grid cells with an L1 distance error below 2.5m. The surface reconstruction accuracy metrics include Chamfer Distance (CD) and F1 \citep{r4}.

\subsection{Generalization Results}
We compare the proposed method with several representative approaches, including PixelSplat \citep{r9}, MVSplat \citep{r24}, DepthSplat \citep{r26}, HiSplat \citep{r27}, Transplat \citep{r25}, and SkySplat \citep{r35}. Since the source code of SkySplat has not been publicly released, we directly report its numerical results from the published paper. The comparison is conducted from multiple perspectives, including the quality of rendered images and height maps, cross-dataset generalization performance, and mesh reconstruction quality.

\subsubsection{Render images and Height maps}
\begin{figure*}
    \centering
    \includegraphics[width=0.9\linewidth]{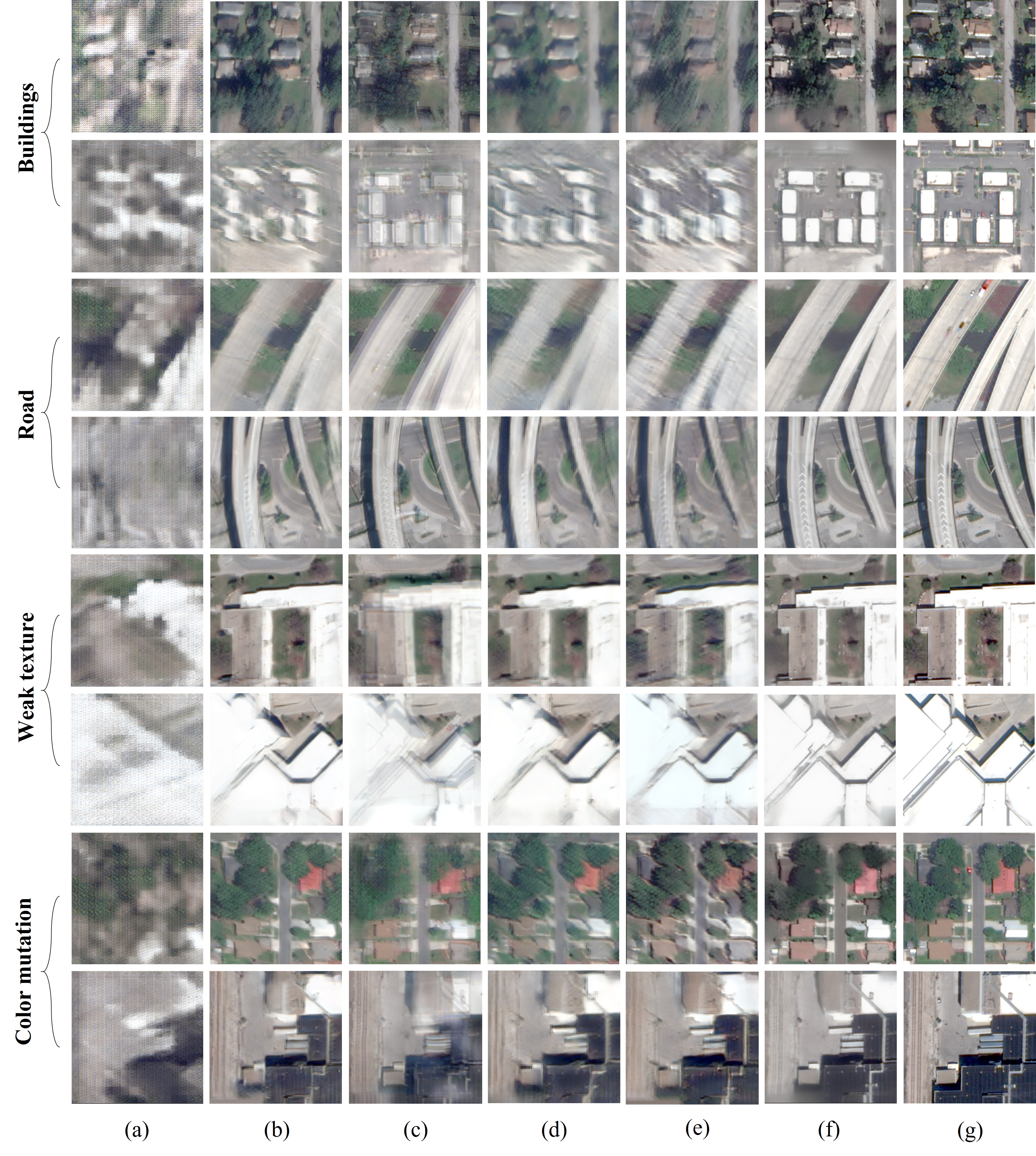}
    \caption{Novel view synthesis results of different methods on the DFC19 dataset. (a) PixelSplat; (b) MVSplat; (c) DepthSplat; (d) HiSplat; (e) Transplat; (f) Ours; (g) Ground truth.}
    \label{fig:renderDFC}
\end{figure*}

\begin{figure*}
    \centering
    \includegraphics[width=0.85\linewidth]{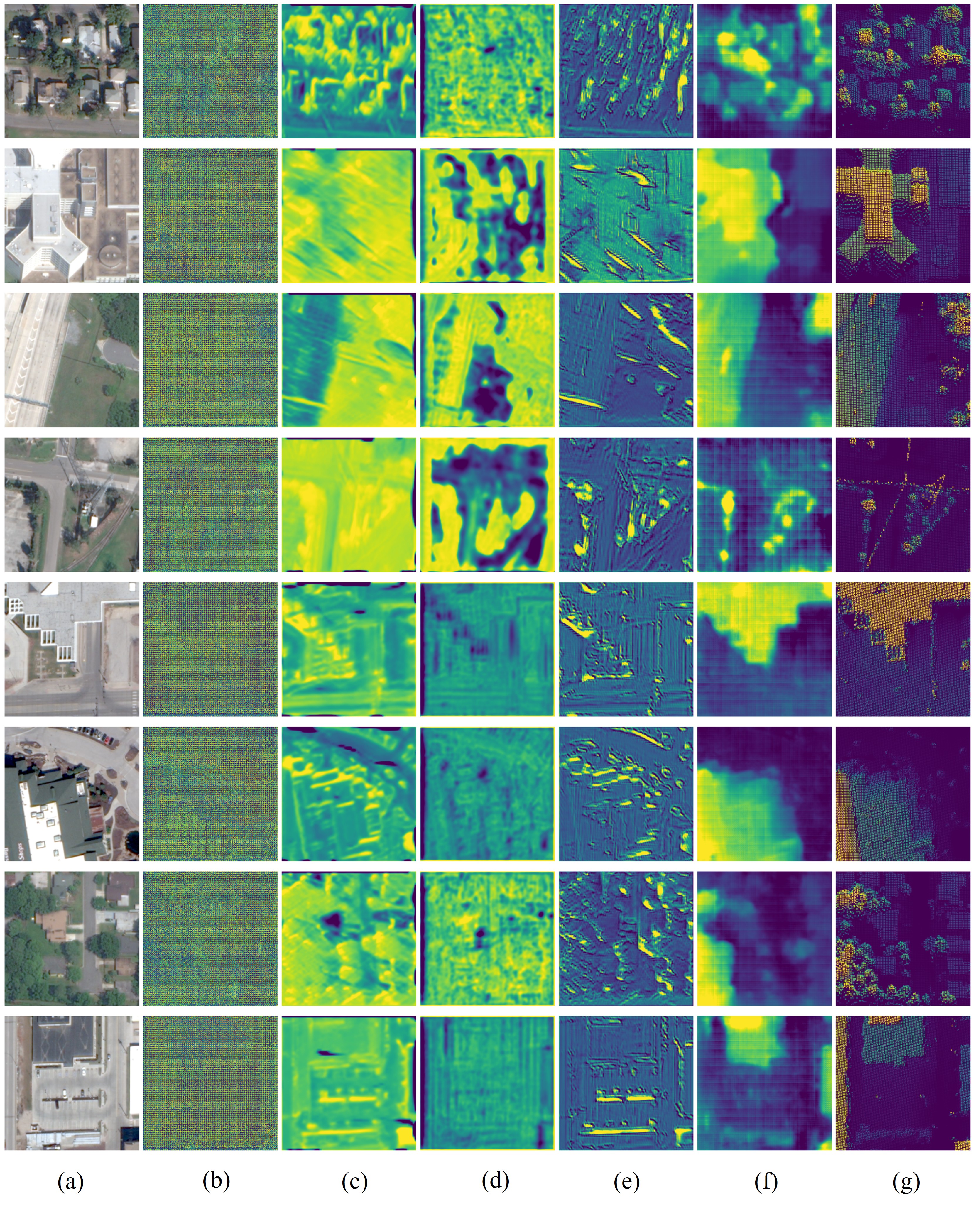}
    \caption{Novel-view height estimation results of different methods on the DFC19 dataset. (a) PixelSplat; (b) MVSplat; (c) DepthSplat; (d) HiSplat; (e) Transplat; (f) Ours; (g) Ground truth.}
    \label{fig:heightmapDFC}
\end{figure*}

We first evaluate the rendering and height estimation performance of different methods on the validation set of the DFC19 dataset. Both quantitative results and visual comparisons are provided to demonstrate the advantages of the proposed method in preserving appearance details and recovering height structures. The quantitative results are presented in TABLE \ref{tab1}, while the visual comparison results are shown in Fig. \ref{fig:renderDFC} and Fig. \ref{fig:heightmapDFC}.

\begin{table*}\rmfamily
\centering
\caption{Quantitative results of appearance reconstruction for novel view synthesis on DFC19 datasets. ``-'' indicates that data is not available. ``Red'', ``Orange'', and ``Yellow'' denote the best, second-best, and third-best results, respectively.}
\label{tab1}
\resizebox{0.9\linewidth}{!}{
\begin{tabular}{lccc|cccc}
\toprule
\multirow{2}{*}{Method} 
& \multicolumn{3}{c|}{Appearance Reconstruction} 
& \multicolumn{4}{c}{Height Estimation} \\
\cmidrule(lr){2-4} \cmidrule(lr){5-8}
& PSNR$\uparrow$ & SSIM$\uparrow$ & LPIPS$\downarrow$ & MAE (m)$\downarrow$ & RMSE (m)$\downarrow$ & PAG$_{2.5}$ (\%)$\uparrow$ & PAG$_{7.5}$ (\%)$\uparrow$ \\
\midrule
PixelSplat \citep{r9}& 16.55 & 0.301 & 0.599 & 13.94 & 17.22 & 19.76 & 37.72 \\
MVSplat \citep{r24}& \secondcell{21.44} & \secondcell{0.590} & \thirdcell{0.415} & 8.39 & 8.92 & 13.51 & 60.61 \\
DepthSplat \citep{r26}& 20.68 & 0.572 & \secondcell{0.412} & 5.65 & 7.05 & 44.88 & 72.65 \\
HiSplat \citep{r27}& 21.19 & 0.580 & 0.443 & 27.91 & 28.37 & 0.58 & 2.88 \\
Transplat \citep{r25}& \thirdcell{21.27} & \thirdcell{0.586} & 0.434 & \thirdcell{4.34} & \thirdcell{5.75} & \thirdcell{51.81} & \thirdcell{79.56} \\
SkySplat \citep{r35}& 18.54 & - & 0.422 & \secondcell{1.80} & \secondcell{2.68} & \secondcell{78.27} & \bestcell{95.57} \\
Ours       & \bestcell{23.68} & \bestcell{0.839} & \bestcell{0.319} & \bestcell{1.72} & \bestcell{2.59} & \bestcell{83.31} & \secondcell{95.41} \\
\bottomrule
\end{tabular}
}
\end{table*}

As shown in the TABLE \ref{tab1}, the proposed method achieves the best overall performance on the DFC19 validation set, demonstrating superior rendering quality and height estimation capability. In terms of appearance reconstruction, our method attains the highest PSNR and SSIM, as well as the lowest LPIPS, indicating that it outperforms all competing methods in overall fidelity, structural consistency, and perceptual quality. Compared with the second-best method, MVSplat, our method further improves PSNR by 2.24 dB, increases SSIM by 0.249, and reduces LPIPS by 0.096. These results suggest that it can not only recover more accurate color information, but also better preserve fine textures and edge structures of the scene. In terms of height estimation, the proposed method also exhibits a clear advantage. It achieves the best MAE and RMSE of 1.72 m and 2.59 m, respectively, among all compared methods. Moreover, it attains the highest $PAG_{2.5}$ of 83.31\% under the stricter accuracy threshold, further demonstrating its ability to recover fine-grained local elevation variations more accurately.

We further present rendering comparisons of different methods over four representative region types, namely Buildings, Road, Weak texture, and Color mutation. As shown in Fig. \ref{fig:renderDFC}, existing methods in satellite scenes are still widely affected by weak textures, repetitive structures, and cross-view photometric variations, and thus tend to suffer from blurring, ghosting, local deformation, and color distortion. In contrast, the proposed method can more clearly recover fine-grained appearance details in building contours, road boundaries, weak-texture regions, and areas with abrupt color changes, while still maintaining strong texture distinguishability and spatial continuity in challenging regions. These results indicate that the proposed method can more effectively distinguish reliable matching regions from degraded ones during multi-view information aggregation, thereby reducing appearance diffusion and detail degradation caused by erroneous correspondences.

The height visualization results further support this observation. As shown in Fig. \ref{fig:heightmapDFC}, compared with the competing methods, which often suffer from blurred boundaries, local noise, fragmented structures, and excessive smoothing, the elevation maps produced by the proposed method exhibit stronger structural integrity and better object separability. They more accurately capture the relative height relationships among buildings, roads, trees, and other ground objects, while appearing particularly more natural around object boundaries and in areas with local terrain undulations. These results indicate that the proposed method not only improves the overall accuracy of elevation estimation, but more importantly enhances the representation of local surface structures, thereby yielding geometric reconstruction results that are more consistent with the true morphology of real-world scenes.

\subsubsection{Cross-Dataset Generalization}
\begin{figure*}
    \centering
    \includegraphics[width=0.85\linewidth]{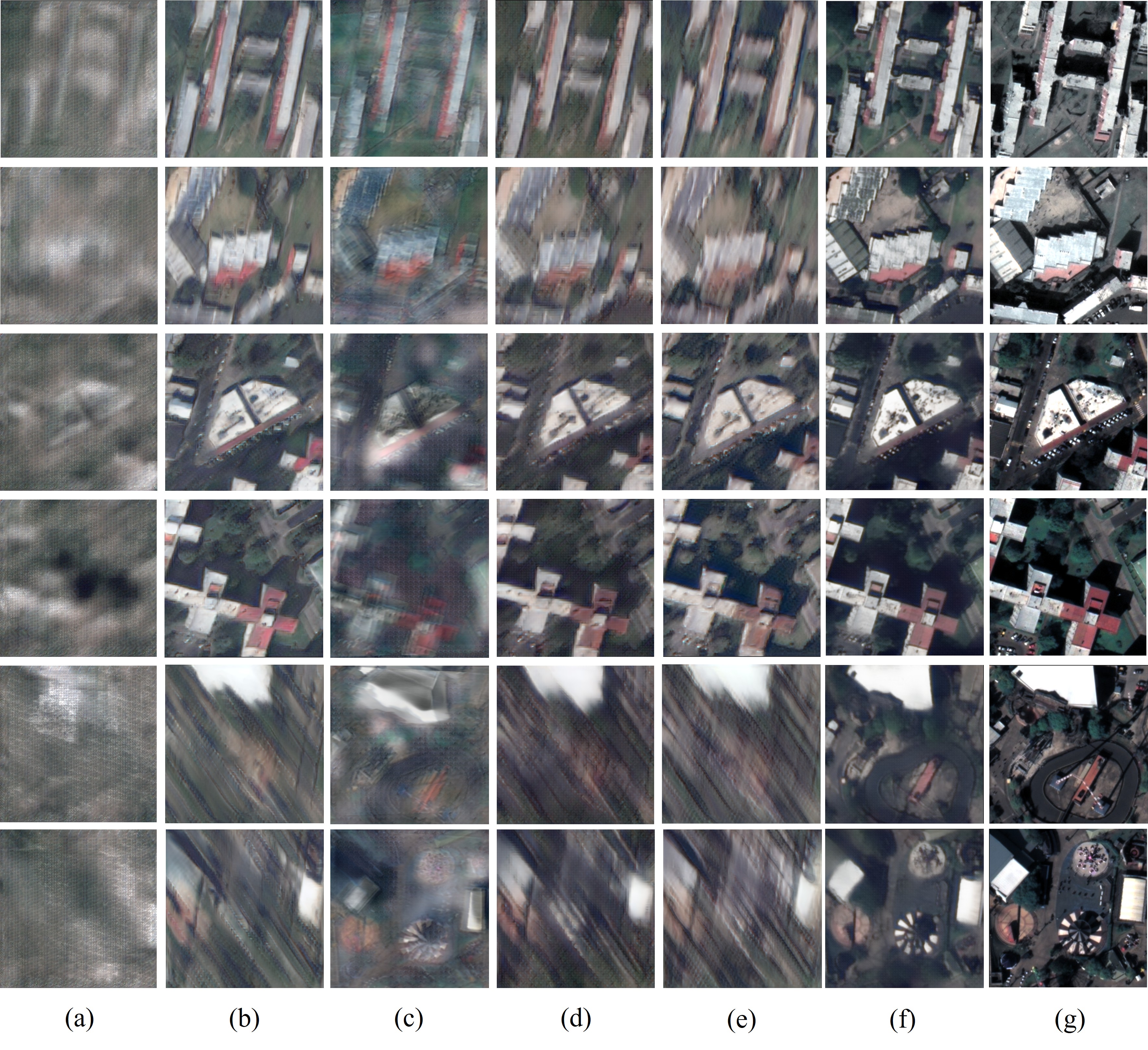}
    \caption{Novel view synthesis results of different methods on the MVS3D dataset. (a) PixelSplat; (b) MVSplat; (c) DepthSplat; (d) HiSplat; (e) Transplat; (f) Ours; (g) Ground truth.}
    \label{fig:renderMVS3D}
\end{figure*}

\begin{figure*}
    \centering
    \includegraphics[width=0.85\linewidth]{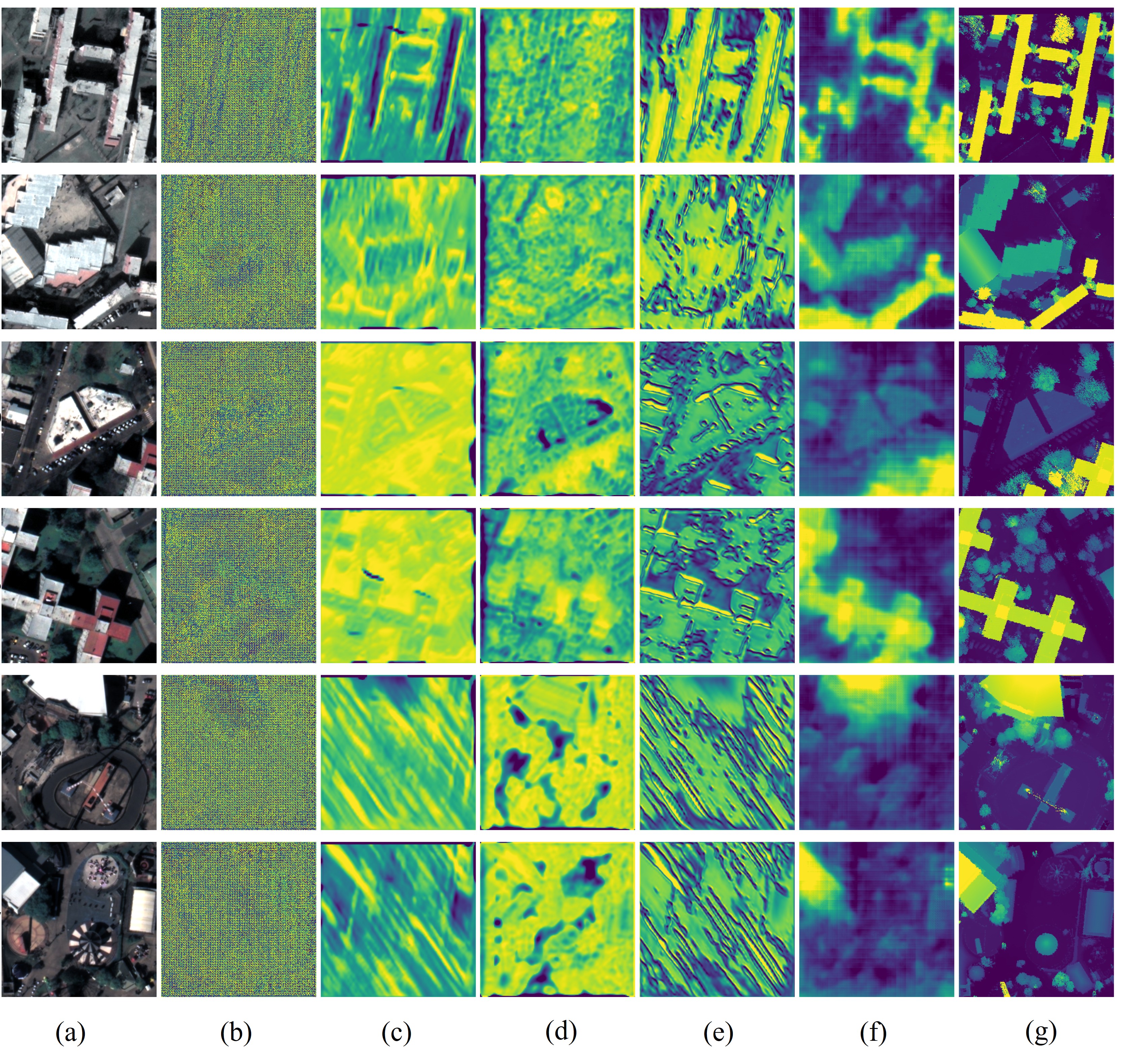}
    \caption{Novel-view height estimation results of different methods on the MVS3D dataset. (a) PixelSplat; (b) MVSplat; (c) DepthSplat; (d) HiSplat; (e) Transplat; (f) Ours; (g) Ground truth.}
    \label{fig:heightmapMVS3D}
\end{figure*}

We further evaluate the cross-dataset generalization performance of the pretrained models on the MVS3D dataset. The quantitative results are presented in TABLE \ref{tab2}, and the corresponding rendering and height estimation results are shown in Fig. \ref{fig:renderMVS3D} and Fig. \ref{fig:heightmapMVS3D}, respectively. As shown in TABLE \ref{tab2}, most competing methods exhibit more pronounced performance degradation under the cross-dataset setting when tested on the unseen MVS3D data. In contrast, the proposed method still achieves the best rendering metrics and delivers superior accuracy in height estimation, demonstrating its ability to maintain strong and stable generalization performance across datasets.

\begin{table*}\rmfamily
\centering
\caption{Quantitative results of appearance reconstruction for novel view synthesis on MVS3D datasets. ``-'' indicates that data is not available. ``Red'', ``Orange'', and ``Yellow'' denote the best, second-best, and third-best results, respectively.}
\label{tab2}
\resizebox{0.85\linewidth}{!}{
\begin{tabular}{lccc|cccc}
\toprule
\multirow{2}{*}{Method}
& \multicolumn{3}{c|}{Appearance Reconstruction}
& \multicolumn{4}{c}{Height Estimation} \\
\cmidrule(lr){2-4} \cmidrule(lr){5-8}
& PSNR$\uparrow$ & SSIM$\uparrow$ & LPIPS$\downarrow$
& MAE (m)$\downarrow$ & RMSE (m)$\downarrow$ & PAG$_{2.5}$ (\%)$\uparrow$ & PAG$_{7.5}$ (\%)$\uparrow$ \\
\midrule
PixelSplat \citep{r9}& 12.14 & 0.218 & 0.632 & 18.38 & 22.81 & \thirdcell{12.65} & 29.79 \\
MVSplat \citep{r24}& \thirdcell{14.58} & \thirdcell{0.370} & \secondcell{0.538} & \thirdcell{8.45} & \thirdcell{9.39} & 10.26 & \thirdcell{46.50} \\
DepthSplat \citep{r26}& 12.58 & 0.261 & 0.612 & 13.92 & 15.25 & 2.35 & 21.36 \\
HiSplat \citep{r27}& \secondcell{16.06} & \secondcell{0.384} & \thirdcell{0.568} & 37.54 & 38.10 & 0.04 & 0.05 \\
Transplat \citep{r25}& 14.43 & 0.346 & 0.577 & 10.64 & 12.19 & 0.47 & 45.01 \\
SkySplat \citep{r35}& - & - & - & \secondcell{3.42} & \bestcell{4.79} & \secondcell{52.32} & \bestcell{89.35} \\
Ours       & \bestcell{17.88} & \bestcell{0.629} & \bestcell{0.468} & \bestcell{3.38} & \secondcell{5.11} & \bestcell{60.90} & \secondcell{86.20} \\
\bottomrule
\end{tabular}
}
\end{table*}

The rendering results in Fig. \ref{fig:renderMVS3D} provide a more intuitive demonstration of this generalization advantage. When faced with more complex building shapes, shadow distributions, and local texture variations in the cross-dataset scenarios, existing methods generally suffer from varying degrees of blurring, structural stretching, edge diffusion, and color drifting, which are particularly evident in slender buildings, sloped roofs, and locally occluded regions. In contrast, the proposed method is able to recover roof outlines, building boundaries, and local structural relationships more completely, while maintaining better detail discriminability and spatial continuity in challenging regions. It also produces more natural shadow transitions and color variations. These observations suggest that the proposed method does not overly rely on the appearance distribution of the training set under cross-domain conditions, but instead is able to maintain plausible rendering results through more stable geometric constraints.

The height estimation results further support this conclusion. As shown in Fig. \ref{fig:heightmapMVS3D}, under the cross-dataset setting, the elevation maps produced by the competing methods are more prone to noise amplification, boundary fragmentation, local misalignment, and excessive smoothing. These issues lead to inconsistent internal heights within building regions, diffused structural contours, and even difficulty in distinguishing the true height differences between adjacent objects. In contrast, the elevation distribution recovered by the proposed method exhibits better structural integrity and object separability. The planar shapes, boundary locations, and local height differences of buildings are represented more clearly, indicating that the proposed method can not only preserve the overall geometric structure but also recover surface shapes more faithfully.

Overall, the cross-dataset experiments indicate that the proposed method is not limited to the statistical characteristics of the source-domain data, but instead learns more transferable geometric and appearance priors for satellite scenes. As a result, it is still able to maintain stable rendering quality and elevation accuracy when confronted with new scene distributions. This also indirectly verifies the effectiveness of the proposed method in integrating multi-view geometry with prior constraints.

\subsubsection{Comparison with per-scene optimization methods}
To validate the runtime advantage of the proposed method, we compare it with two representative per-scene optimization methods, namely EOGS \citep{r8} and Sat-nerf \citep{r7}. For a fair comparison, we set the number of training views for EOGS and Sat-nerf to three and use one additional view for validation, following the same setting as our method. Both EOGS and Sat-nerf were trained for 10000 iterations.

As shown in TABLE \ref{tab3}, compared with two representative per-scene optimization methods, the proposed method achieves a better overall balance between efficiency and performance. Although EOGS and Sat-NeRF allow repeated iterative optimization for each individual scene, under the setting of using only three training views and a unified budget of 10000 iterations, our method still attains better rendering quality on average and significantly more accurate height estimation, while requiring only seconds for inference. In contrast, per-scene optimization methods typically require hundreds to thousands of seconds to complete the reconstruction of a single scene, resulting in substantially higher computational cost. These results indicate that the advantage of the proposed method is not merely due to faster feed-forward inference; rather, by learning cross-scene priors, it can directly produce competitive or even superior reconstruction results with extremely low time overhead.

From the overall trend, per-scene optimization methods still retain certain advantages in some appearance metrics for individual scenes, reflecting their ability to achieve more thorough scene-specific fitting of local photometric distributions through prolonged optimization. However, this advantage in appearance fitting does not consistently translate into more reliable geometric recovery. In most scenes, these methods still exhibit relatively large height errors, and their accuracy coverage under both threshold settings remains noticeably lower. In contrast, the proposed method demonstrates more stable appearance–geometry consistency across different scenes and image resolutions.

\begin{table*}[ht]\rmfamily
\centering
\caption{Quantitative comparison of rendering quality, geometric accuracy, and runtime between the proposed method with the per-scene optimization methods Sat-nerf \citep{r7} and EOGS \citep{r8} across different scenes. ``Red'' and ``Orange'' denote the best and second-best results, respectively.}
\label{tab3}
\resizebox{\textwidth}{!}{
\begin{tabular}{llccc|cccc|c}
\toprule
\multirow{2}{*}{Scene} & \multirow{2}{*}{Method}
& \multicolumn{3}{c|}{Appearance Reconstruction}
& \multicolumn{4}{c|}{Height Estimation}
& \multirow{2}{*}{Time (s)$\downarrow$} \\
\cmidrule(lr){3-5} \cmidrule(lr){6-9}
& & PSNR$\uparrow$ & SSIM$\uparrow$ & LPIPS$\downarrow$
& MAE (m)$\downarrow$ & RMSE (m)$\downarrow$ & PAG$_{2.5}$ (\%)$\uparrow$ & PAG$_{7.5}$ (\%)$\uparrow$ & \\
\midrule

\multirow{3}{*}{Mean}
& Sat-nerf & 20.26 & \secondcell{0.536} & 0.561 & \secondcell{11.453} & \secondcell{14.110} & \secondcell{16.11} & \secondcell{45.68} & 1229.15 \\
& EOGS     & \secondcell{20.76} & 0.526 & \secondcell{0.412} & 18.576 & 23.020 & 15.94 & 35.04 & \secondcell{463.70} \\
& Ours     & \bestcell{21.11} & \bestcell{0.759} & \bestcell{0.379} & \bestcell{2.760} & \bestcell{4.455} & \bestcell{72.51} & \bestcell{89.98} & \bestcell{1.58} \\
\midrule

\multirow{3}{*}{\shortstack[l]{IAPRA001\\(736$\times$736)}}
& Sat-nerf & \bestcell{19.78} & 0.544 & 0.544 & 10.59 & 13.02 & 14.77 & 42.33 & 1662.80 \\
& EOGS     & \secondcell{19.44} & \secondcell{0.560} & \bestcell{0.400} & \secondcell{7.89} & \secondcell{10.06} & \secondcell{25.61} & \secondcell{48.49} & \secondcell{310.41} \\
& Ours     & 15.72 & \bestcell{0.594} & \secondcell{0.459} & \bestcell{2.732} & \bestcell{4.122} & \bestcell{66.01} & \bestcell{90.19} & \bestcell{1.37} \\
\midrule

\multirow{3}{*}{\shortstack[l]{IAPRA002\\(736$\times$736)}}
& Sat-nerf & \secondcell{16.50} & \secondcell{0.381} & 0.647 & 16.44 & 20.27 & 9.47 & 27.34 & 1134.82 \\
& EOGS     & 16.09 & 0.353 & \secondcell{0.562} & \secondcell{15.47} & \secondcell{20.14} & \secondcell{14.94} & \secondcell{31.12} & \secondcell{316.14} \\
& Ours     & \bestcell{19.62} & \bestcell{0.691} & \bestcell{0.420} & \bestcell{3.956} & \bestcell{6.023} & \bestcell{57.19} & \bestcell{80.89} & \bestcell{1.08} \\
\midrule

\multirow{3}{*}{\shortstack[l]{IAPRA003\\(736$\times$736)}}
& Sat-nerf & \bestcell{22.72} & \secondcell{0.585} & \secondcell{0.526} & \secondcell{14.56} & \secondcell{17.95} & \secondcell{11.18} & \secondcell{30.43} & 1159.87 \\
& EOGS     & 17.06 & 0.263 & \bestcell{0.486} & 30.21 & 37.55 & 5.92 & 16.60 & \secondcell{311.79} \\
& Ours     & \secondcell{17.77} & \bestcell{0.601} & 0.536 & \bestcell{3.523} & \bestcell{5.394} & \bestcell{62.08} & \bestcell{86.69} & \bestcell{0.95} \\
\midrule

\multirow{3}{*}{\shortstack[l]{JAX004\\(1024$\times$1024)}}
& Sat-nerf & 21.49 & 0.631 & 0.547 & \secondcell{4.23} & \secondcell{5.09} & \secondcell{31.91} & \secondcell{88.10} & 1142.92 \\
& EOGS     & \bestcell{26.27} & \secondcell{0.767} & \bestcell{0.334} & 6.01 & 7.11 & 21.38 & 67.50 & \secondcell{385.14} \\
& Ours     & \secondcell{24.29} & \bestcell{0.824} & \secondcell{0.398} & \bestcell{1.721} & \bestcell{2.719} & \bestcell{79.71} & \bestcell{95.42} & \bestcell{2.24} \\
\midrule

\multirow{3}{*}{\shortstack[l]{JAX068\\(1024$\times$1024)}}
& Sat-nerf & 20.55 & 0.505 & 0.590 & 8.73 & \secondcell{11.07} & 18.08 & 51.08 & 1150.66 \\
& EOGS     & \bestcell{27.03} & \secondcell{0.818} & \bestcell{0.222} & \secondcell{8.42} & 12.21 & \secondcell{35.61} & \secondcell{58.64} & \secondcell{296.27} \\
& Ours     & \secondcell{22.05} & \bestcell{0.835} & \secondcell{0.345} & \bestcell{2.11} & \bestcell{3.93} & \bestcell{80.30} & \bestcell{94.71} & \bestcell{1.80} \\
\midrule

\multirow{3}{*}{\shortstack[l]{JAX214\\(1024$\times$1024)}}
& Sat-nerf & \secondcell{18.78} & \secondcell{0.504} & 0.587 & \secondcell{18.44} & \secondcell{22.66} & \secondcell{8.33} & \secondcell{24.40} & 1176.75 \\
& EOGS     & 14.80 & 0.252 & \secondcell{0.526} & 41.47 & 51.26 & 4.02 & 11.64 & \secondcell{826.72} \\
& Ours     & \bestcell{22.97} & \bestcell{0.856} & \bestcell{0.289} & \bestcell{3.35} & \bestcell{5.83} & \bestcell{79.80} & \bestcell{89.52} & \bestcell{1.79} \\
\midrule

\multirow{3}{*}{\shortstack[l]{JAX260\\(1024$\times$1024)}}
& Sat-nerf & 21.99 & \secondcell{0.599} & 0.489 & \secondcell{7.18} & \secondcell{8.71} & \secondcell{19.03} & \secondcell{56.07} & 1176.24 \\
& EOGS     & \secondcell{24.61} & 0.669 & \secondcell{0.356} & 20.56 & 22.81 & 4.10 & 11.28 & \secondcell{799.43} \\
& Ours     & \bestcell{25.37} & \bestcell{0.912} & \bestcell{0.209} & \bestcell{1.93} & \bestcell{3.17} & \bestcell{82.51} & \bestcell{92.47} & \bestcell{1.82} \\
\bottomrule
\end{tabular}
}
\end{table*}

\begin{figure*}
    \centering
    \includegraphics[width=0.9\linewidth]{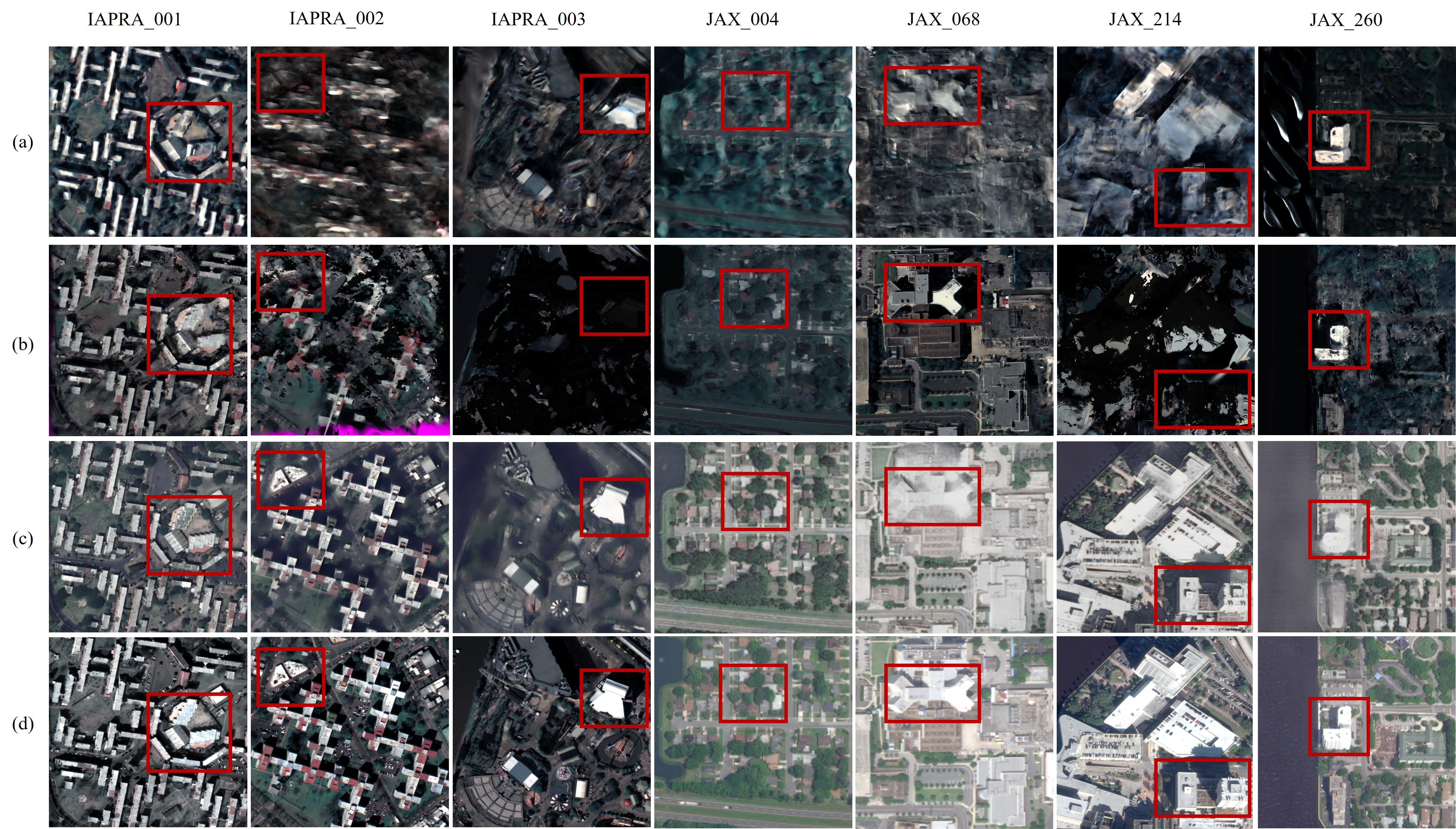}
    \caption{Comparison of novel-view synthesis results between the proposed method and per-scene optimization methods across different scenes. (a) Sat-NeRF; (b) EOGS; (c) Ours; (d) Ground truth.}
    \label{fig:renderOri}
\end{figure*}

\begin{figure*}
    \centering
    \includegraphics[width=0.9\linewidth]{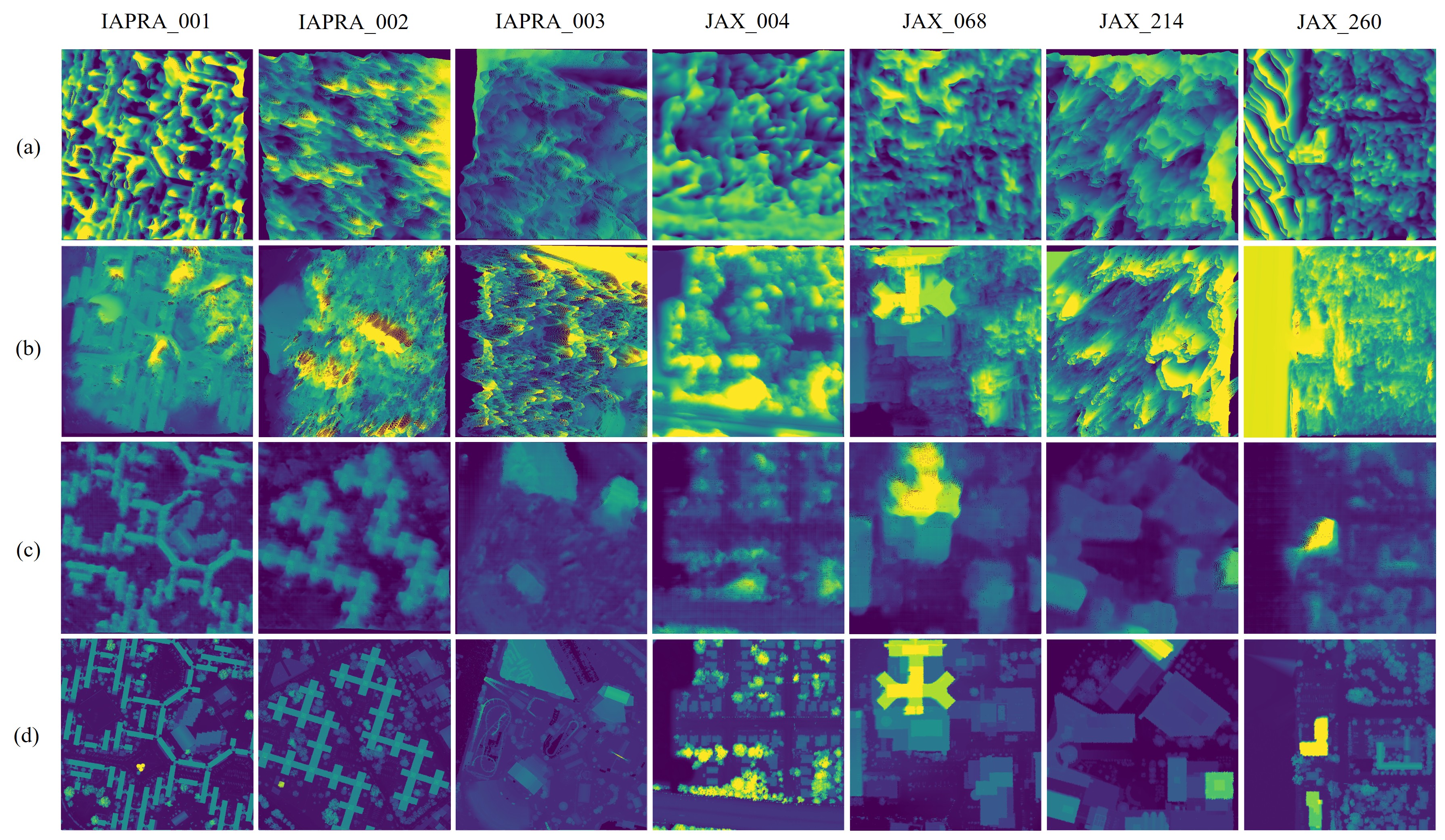}
    \caption{Comparison of novel-view height map estimation results between the proposed method and per-scene optimization methods across different scenes. (a) Sat-NeRF; (b) EOGS; (c) Ours; (d) Ground truth.}
    \label{fig:heightmapOri}
\end{figure*}

The rendering results shown in Fig. \ref{fig:renderOri} further support the above conclusion. Sat-NeRF is able to recover relatively plausible overall brightness and coarse contours in several scenes, but it is often accompanied by noticeable blurring, collapsed shadow regions, or missing local details. EOGS can improve local contrast in some regions, yet it is prone to color drifting, texture distortion, and structural discontinuity. In contrast, in the regions highlighted by the red boxes, the proposed method generally preserves building contours, roof boundaries, and the spatial relationships between adjacent objects more completely, while maintaining better local discriminability and overall visual consistency under complex shadows, dense buildings, and weak-texture backgrounds. These results indicate that, under sparse-input conditions, the proposed method does not overly rely on scene-specific appearance fitting, but is instead able to generate more stable rendering results by leveraging learned geometric and photometric priors.

The height estimation results shown in Fig. \ref{fig:heightmapOri} make the advantage of the proposed method even more evident. The predictions of Sat-NeRF often exhibit strong local fluctuation noise, resulting in blurred building boundaries and inconsistent internal height distributions. EOGS, by contrast, is more prone to excessive smoothing, which noticeably blurs building outlines and object boundaries and makes it difficult to recover clear target shapes. In comparison, the elevation distributions recovered by the proposed method exhibit better structural integrity and object separability. They not only delineate building footprints and relative height relationships more accurately, but also preserve more natural hierarchical transitions in complex urban scenes. These results indicate that the priors learned by the proposed method are not limited to improving rendering appearance, but also enhance the network’s ability to constrain the true geometric structure of the scene.

\subsubsection{Surface Reconstruction}
To further evaluate the surface reconstruction capability of different methods, we compare the quality of the mesh models extracted from each method on the DFC19 and MVS3D datasets. For fairness, we use Truncated Signed Distance Function(TSDF) \citep{r46} extraction pipeline for all methods to generate the mesh models. The quantitative results are reported in TABLE \ref{tab4}. For both datasets, the mesh model generated from all views for each scene was used as the ground truth.

As shown in the TABLE \ref{tab4}, the proposed method achieves the lowest CD and the highest F1 on both datasets, indicating that it attains a better balance between geometric accuracy and surface completeness of the reconstructed meshes. Compared with the second-best method, HiSplat, our method reduces CD by 43\% and improves F1 by 12\% on DFC19. On the more challenging MVS3D dataset, this advantage becomes even more pronounced, with CD reduced by 59.7\% and F1 improved by 59.2\%.

\begin{figure*}
    \centering
    \includegraphics[width=0.75\linewidth]{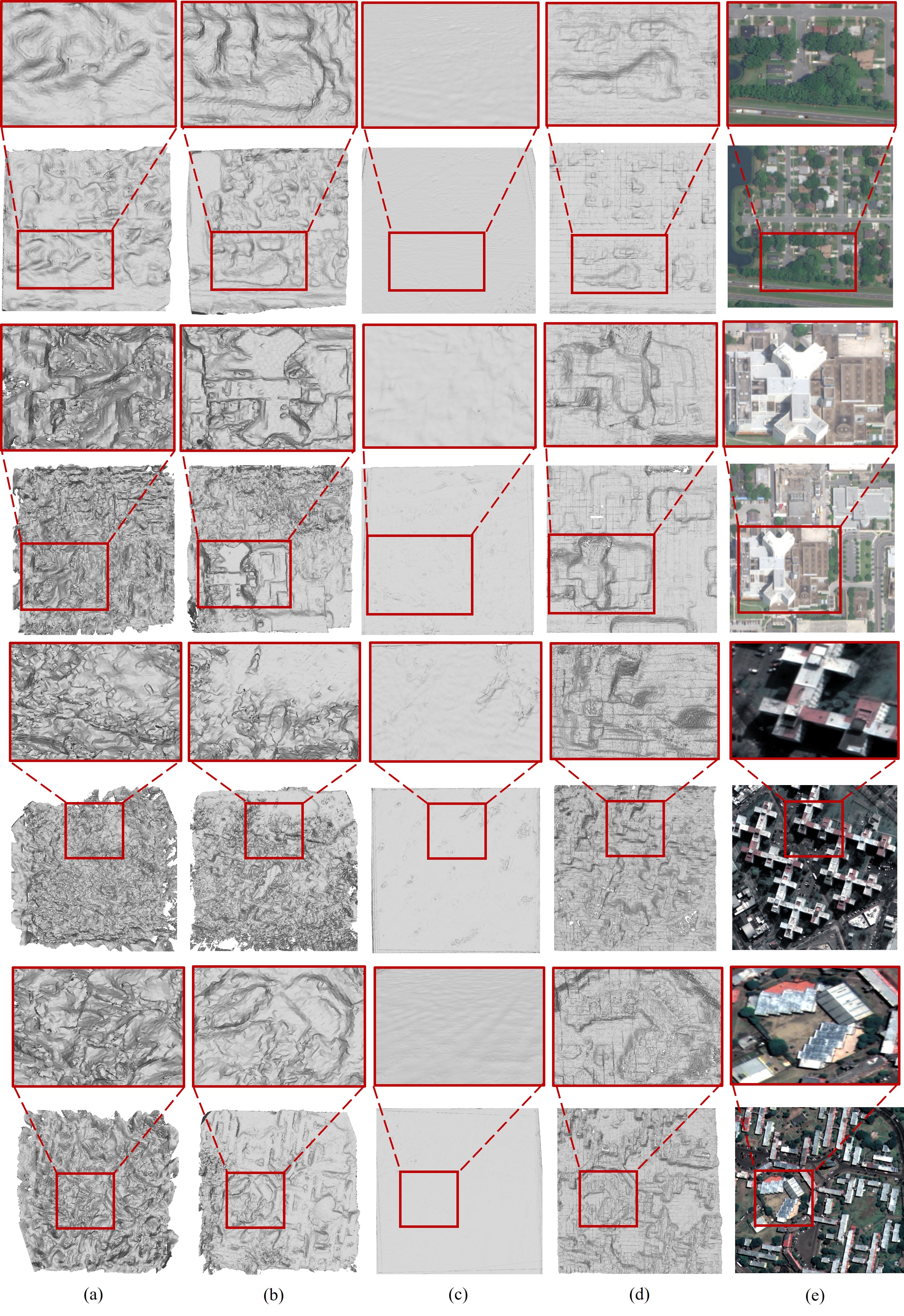}
    \caption{Qualitative comparison of mesh models extracted by different methods. (a) Sat-NeRF; (b) EOGS; (c) HiSplat; (d) Ours; (e) corresponding Image.}
    \label{fig:mesh}
\end{figure*}

The qualitative results shown in Fig. \ref{fig:mesh} further support this conclusion. Compared with the competing methods, the meshes extracted by the proposed method exhibit clearer structural boundaries and more complete object shapes in the regions highlighted by the red boxes. In areas with well-defined geometric contours, such as building roofs and regular man-made facilities, our method is able to more faithfully recover their planar structures and spatial layouts, whereas the other methods often suffer from local surface fragmentation, detail loss, or large areas of spurious undulations. Overall, these results indicate that, under sparse-view satellite scenarios, the proposed method not only improves rendering quality, but more importantly strengthens the ability of the surface representation to constrain the true scene geometry. As a result, it achieves higher accuracy, better completeness, and more stable cross-scene performance in the final mesh reconstruction.

\begin{table}\rmfamily
\centering
\caption{Quantitative comparison of the mesh models extracted by different methods in terms of CD and F1 on the DFC19 and MVS3D datasets. ``Red'', ``Orange'', and ``Yellow'' denote the best, second-best, and third-best results, respectively.}
\label{tab4}
\begin{tabular}{lcc|cc}
\toprule
\multirow{2}{*}{Method} & \multicolumn{2}{c|}{DFC19} & \multicolumn{2}{c}{MVS3D} \\
\cmidrule(lr){2-3} \cmidrule(lr){4-5}
& CD$\downarrow$ & F1$\uparrow$ & CD$\downarrow$ & F1$\uparrow$ \\
\midrule
Satnerf \citep{r7}& 4.71 & 0.13 & 4.43 & 0.16 \\
EOGS \citep{r8}& 4.62 & 0.14 & 4.99 & 0.17 \\
PixelSplat \citep{r9}& 17.19 & 0.09 & 6.52 & 0.13 \\
MVSplat \citep{r24}& 2.72 & 0.46 & 4.11 & \secondcell{0.22} \\
DepthSplat \citep{r26}& \thirdcell{2.69} & 0.43 & 4.15 & \thirdcell{0.21} \\
HiSplat \citep{r27}& \secondcell{2.67} & \secondcell{0.49} & \secondcell{3.95} & \secondcell{0.22} \\
Transplat \citep{r25}& 2.70 & \thirdcell{0.47} & \thirdcell{4.01} & 0.19 \\
Ours       & \bestcell{1.52} & \bestcell{0.55} & \bestcell{1.59} & \bestcell{0.54} \\
\bottomrule
\end{tabular}
\end{table}

\subsection{Ablations}
To validate the effectiveness of the proposed module designs, we conduct ablation experiments on the DFC19 validation set by progressively removing or replacing different components from the full model, thereby analyzing the contribution of each part. Specifically, Naive Fusion denotes a simple baseline that replaces the proposed CMMF and CSRG by directly concatenating monocular and multi-view features followed by a $1\times1$ convolution, without confidence-aware adaptive routing or cross-stage residual guidance. In addition, w/o CBRL denotes removing the proposed confidence bi-directional routing loss while keeping the network architecture and all other training settings unchanged.

\begin{table*}[t]\rmfamily
\centering
\caption{Ablation study on the DFC19 dataset. Naive Fusion denotes a simple baseline that replaces the proposed CMMF and CSRG by directly concatenating monocular and multi-view features followed by a $1\times1$ convolution, without confidence-aware adaptive routing or cross-stage residual guidance. ``Red'', ``Orange'', and ``Yellow'' denote the best, second-best, and third-best results, respectively.}
\label{tab5}
\resizebox{0.9\textwidth}{!}{
\begin{tabular}{lccccccc}
\toprule
Method & PSNR$\uparrow$ & SSIM$\uparrow$ & LPIPS$\downarrow$ & MAE (m)$\downarrow$ & RMSE (m)$\downarrow$ & PAG$_{2.5}$ (\%)$\uparrow$ & PAG$_{7.5}$ (\%)$\uparrow$ \\
\midrule
w/o CMMF\&CSRG & 22.46 & 0.820 & 0.352 & 2.74 & 3.72 & 67.44 & 93.66 \\
Naive Fusion   & 22.67 & 0.827 & 0.355 & 1.91 & 3.09 & 80.41 & 94.65 \\
w/o CSRG       & \thirdcell{22.92} & \thirdcell{0.829} & \thirdcell{0.332} & \thirdcell{1.86} & \thirdcell{3.01} & \thirdcell{81.70} & \secondcell{94.87} \\
w/o CMMF       & \secondcell{23.29} & \secondcell{0.836} & \secondcell{0.325} & 2.27 & 3.30 & 76.93 & 94.47 \\
w/o CBRL       & 22.71 & 0.827 & 0.337 & \secondcell{1.82} & \secondcell{2.97} & \secondcell{81.83} & \thirdcell{94.72} \\
\midrule
Full model     & \bestcell{23.68} & \bestcell{0.839} & \bestcell{0.319} & \bestcell{1.72} & \bestcell{2.59} & \bestcell{83.31} & \bestcell{95.41} \\
\bottomrule
\end{tabular}
}
\end{table*}

The results are reported in Table \ref{tab5}. Overall, the full model achieves the best performance across all metrics, demonstrating that the proposed components work collaboratively to improve both rendering quality and height estimation. Compared with the variants obtained by removing different components, the full model shows stronger appearance reconstruction capability in terms of PSNR,SSIM and LPIPS , while also achieving the best results in MAE,RMSE , $PAG_{2.5}$ and $PAG_{7.5}$. These findings indicate that the proposed method can simultaneously enhance photometric consistency and geometric recovery accuracy.

First, when CMMF and CSRG are removed simultaneously, the model exhibits the most pronounced performance degradation, indicating that these two core modules are essential to the overall performance improvement. Furthermore, when Naive Fusion is used to replace CMMF and CSRG, the performance shows a certain degree of recovery compared with directly removing both modules, with the improvement being particularly more evident on the geometric metrics. This suggests that cross-source interaction between monocular priors and multi-view features is itself beneficial. However, Naive Fusion still remains clearly inferior to the full model, indicating that the performance gain does not arise merely from simple feature fusion, but mainly from the proposed confidence-aware adaptive fusion and cross-stage residual guidance mechanisms.

From the single-module ablation results, removing CMMF leads to more pronounced degradation in the height estimation metrics, particularly in MAE, RMSE, and $PAG_{2.5}$ This indicates that CMMF can more effectively coordinate monocular priors and multi-view geometric information in regions with weak textures, repetitive textures, and degraded multi-view matching, thereby improving the accuracy of local geometric recovery. By contrast, removing CSRG results in declines in both appearance and geometric metrics, but with a relatively more balanced degradation pattern. This suggests that the module primarily improves the consistency of surface representation and the stability of optimization by introducing explicit cues of cross-stage geometric inconsistency to stabilize the stage-wise update of Gaussian parameters. Furthermore, removing CBRL results in a decline in model performance, indicating that the proposed confidence-based bidirectional branching loss effectively improves the rationality of supervision allocation during training.

\section{Conclusion}
In this paper, we presented SatSurfGS, a feed-forward 2D Gaussian Splatting framework for sparse-view satellite surface reconstruction under RPC camera model. Geometric reliability in satellite imagery is spatially heterogeneous due to large photometric variations, weak or repetitive textures, and unstable multi-view matching. Based on this perspective, SatSurfGS explicitly models local reliability throughout feature learning, Gaussian parameter estimation, and training optimization. Specifically, CMMF enables adaptive cross-source information selection between monocular priors and multiview cues, CSRG stabilizes stage-wise Gaussian refinement by introducing confidence-aware cross-stage geometric inconsistency guidance, and CBRL further improves training by routing geometric and appearance supervision according to local reliability. Extensive experiments on both in-domain and cross-dataset benchmarks demonstrate that SatSurfGS achieves superior rendering quality, more accurate height recovery, and better surface reconstruction performance, while maintaining clear efficiency advantages over competitive per-scene optimization methods. These results verify that reliability-aware modeling provides an effective way to build a generalizable 2DGS framework for sparse-view satellite surface reconstruction.

\textbf{Limitations and Future Work.} Current limitations include the current study mainly focuses on static sparse-view satellite scenes, and the modeling of more complex large-scale or temporally varying scenes remains underexplored. In future work, we plan to investigate stronger geometry-aware priors, more robust uncertainty modeling, and more scalable reconstruction strategies for large-area satellite scenes. We are also interested in extending the framework toward more accurate surface-oriented representations and more challenging real-world settings, such as multi-temporal reconstruction and broader cross-sensor generalization.

\printcredits

\bibliographystyle{cas-model2-names}

\bibliography{cas-refs}

\begin{thebibliography}{45}
\expandafter\ifx\csname natexlab\endcsname\relax\def\natexlab#1{#1}\fi
\providecommand{\url}[1]{\texttt{#1}}
\providecommand{\href}[2]{#2}
\providecommand{\path}[1]{#1}
\providecommand{\DOIprefix}{doi:}
\providecommand{\ArXivprefix}{arXiv:}
\providecommand{\URLprefix}{URL: }
\providecommand{\Pubmedprefix}{pmid:}
\providecommand{\doi}[1]{\href{http://dx.doi.org/#1}{\path{#1}}}
\providecommand{\Pubmed}[1]{\href{pmid:#1}{\path{#1}}}
\providecommand{\bibinfo}[2]{#2}
\ifx\xfnm\relax \def\xfnm[#1]{\unskip,\space#1}\fi
\bibitem[{Aira et~al.(2025)Aira, Facciolo and Ehret}]{r8}
\bibinfo{author}{Aira, L.S.}, \bibinfo{author}{Facciolo, G.}, \bibinfo{author}{Ehret, T.}, \bibinfo{year}{2025}.
\newblock \bibinfo{title}{Gaussian splatting for efficient satellite image photogrammetry}, in: \bibinfo{booktitle}{Proceedings of the Computer Vision and Pattern Recognition Conference}, pp. \bibinfo{pages}{5959--5969}.
\bibitem[{Bai et~al.(2025)Bai, Yang, Chen and Du}]{r34}
\bibinfo{author}{Bai, N.}, \bibinfo{author}{Yang, A.}, \bibinfo{author}{Chen, H.}, \bibinfo{author}{Du, C.}, \bibinfo{year}{2025}.
\newblock \bibinfo{title}{Satgs: Remote sensing novel view synthesis using multi-temporal satellite images with appearance-adaptive 3dgs}.
\newblock \bibinfo{journal}{Remote Sensing} \bibinfo{volume}{17}, \bibinfo{pages}{1609}.
\bibitem[{Barron et~al.(2022)Barron, Mildenhall, Verbin, Srinivasan and Hedman}]{r11}
\bibinfo{author}{Barron, J.T.}, \bibinfo{author}{Mildenhall, B.}, \bibinfo{author}{Verbin, D.}, \bibinfo{author}{Srinivasan, P.P.}, \bibinfo{author}{Hedman, P.}, \bibinfo{year}{2022}.
\newblock \bibinfo{title}{Mip-nerf 360: Unbounded anti-aliased neural radiance fields}, in: \bibinfo{booktitle}{Proceedings of the IEEE/CVF conference on computer vision and pattern recognition}, pp. \bibinfo{pages}{5470--5479}.
\bibitem[{Barron et~al.(2023)Barron, Mildenhall, Verbin, Srinivasan and Hedman}]{r10}
\bibinfo{author}{Barron, J.T.}, \bibinfo{author}{Mildenhall, B.}, \bibinfo{author}{Verbin, D.}, \bibinfo{author}{Srinivasan, P.P.}, \bibinfo{author}{Hedman, P.}, \bibinfo{year}{2023}.
\newblock \bibinfo{title}{Zip-nerf: Anti-aliased grid-based neural radiance fields}, in: \bibinfo{booktitle}{Proceedings of the IEEE/CVF International Conference on Computer Vision}, pp. \bibinfo{pages}{19697--19705}.
\bibitem[{Bosch et~al.(2019)Bosch, Foster, Christie, Wang, Hager and Brown}]{r44}
\bibinfo{author}{Bosch, M.}, \bibinfo{author}{Foster, K.}, \bibinfo{author}{Christie, G.}, \bibinfo{author}{Wang, S.}, \bibinfo{author}{Hager, G.D.}, \bibinfo{author}{Brown, M.}, \bibinfo{year}{2019}.
\newblock \bibinfo{title}{Semantic stereo for incidental satellite images}, in: \bibinfo{booktitle}{2019 IEEE Winter Conference on Applications of Computer Vision (WACV)}, \bibinfo{organization}{IEEE}. pp. \bibinfo{pages}{1524--1532}.
\bibitem[{Bosch et~al.(2016)Bosch, Kurtz, Hagstrom and Brown}]{r45}
\bibinfo{author}{Bosch, M.}, \bibinfo{author}{Kurtz, Z.}, \bibinfo{author}{Hagstrom, S.}, \bibinfo{author}{Brown, M.}, \bibinfo{year}{2016}.
\newblock \bibinfo{title}{A multiple view stereo benchmark for satellite imagery}, in: \bibinfo{booktitle}{2016 IEEE Applied Imagery Pattern Recognition Workshop (AIPR)}, \bibinfo{organization}{IEEE}. pp. \bibinfo{pages}{1--9}.
\bibitem[{Charatan et~al.(2024)Charatan, Li, Tagliasacchi and Sitzmann}]{r9}
\bibinfo{author}{Charatan, D.}, \bibinfo{author}{Li, S.L.}, \bibinfo{author}{Tagliasacchi, A.}, \bibinfo{author}{Sitzmann, V.}, \bibinfo{year}{2024}.
\newblock \bibinfo{title}{pixelsplat: 3d gaussian splats from image pairs for scalable generalizable 3d reconstruction}, in: \bibinfo{booktitle}{Proceedings of the IEEE/CVF conference on computer vision and pattern recognition}, pp. \bibinfo{pages}{19457--19467}.
\bibitem[{Chen et~al.(2024)Chen, Xu, Zheng, Zhuang, Pollefeys, Geiger, Cham and Cai}]{r24}
\bibinfo{author}{Chen, Y.}, \bibinfo{author}{Xu, H.}, \bibinfo{author}{Zheng, C.}, \bibinfo{author}{Zhuang, B.}, \bibinfo{author}{Pollefeys, M.}, \bibinfo{author}{Geiger, A.}, \bibinfo{author}{Cham, T.J.}, \bibinfo{author}{Cai, J.}, \bibinfo{year}{2024}.
\newblock \bibinfo{title}{Mvsplat: Efficient 3d gaussian splatting from sparse multi-view images}, in: \bibinfo{booktitle}{European conference on computer vision}, \bibinfo{organization}{Springer}. pp. \bibinfo{pages}{370--386}.
\bibitem[{Chen et~al.(2025)Chen, Wu, Shen, Zhao, Ye, Feng, Ding and Zhang}]{r28}
\bibinfo{author}{Chen, Z.}, \bibinfo{author}{Wu, C.}, \bibinfo{author}{Shen, Z.}, \bibinfo{author}{Zhao, C.}, \bibinfo{author}{Ye, W.}, \bibinfo{author}{Feng, H.}, \bibinfo{author}{Ding, E.}, \bibinfo{author}{Zhang, S.H.}, \bibinfo{year}{2025}.
\newblock \bibinfo{title}{Splatter-360: Generalizable 360 gaussian splatting for wide-baseline panoramic images}, in: \bibinfo{booktitle}{Proceedings of the Computer Vision and Pattern Recognition Conference}, pp. \bibinfo{pages}{21590--21599}.
\bibitem[{Cheng et~al.(2020)Cheng, Xu, Zhu, Li, Li, Ramamoorthi and Su}]{r38}
\bibinfo{author}{Cheng, S.}, \bibinfo{author}{Xu, Z.}, \bibinfo{author}{Zhu, S.}, \bibinfo{author}{Li, Z.}, \bibinfo{author}{Li, L.E.}, \bibinfo{author}{Ramamoorthi, R.}, \bibinfo{author}{Su, H.}, \bibinfo{year}{2020}.
\newblock \bibinfo{title}{Deep stereo using adaptive thin volume representation with uncertainty awareness}, in: \bibinfo{booktitle}{Proceedings of the IEEE/CVF conference on computer vision and pattern recognition}, pp. \bibinfo{pages}{2524--2534}.
\bibitem[{Gao et~al.(2023)Gao, Liu and Ji}]{r37}
\bibinfo{author}{Gao, J.}, \bibinfo{author}{Liu, J.}, \bibinfo{author}{Ji, S.}, \bibinfo{year}{2023}.
\newblock \bibinfo{title}{A general deep learning based framework for 3d reconstruction from multi-view stereo satellite images}.
\newblock \bibinfo{journal}{ISPRS Journal of Photogrammetry and Remote Sensing} \bibinfo{volume}{195}, \bibinfo{pages}{446--461}.
\bibitem[{Gu{\'e}don and Lepetit(2024)}]{r20}
\bibinfo{author}{Gu{\'e}don, A.}, \bibinfo{author}{Lepetit, V.}, \bibinfo{year}{2024}.
\newblock \bibinfo{title}{Sugar: Surface-aligned gaussian splatting for efficient 3d mesh reconstruction and high-quality mesh rendering}, in: \bibinfo{booktitle}{Proceedings of the IEEE/CVF conference on computer vision and pattern recognition}, pp. \bibinfo{pages}{5354--5363}.
\bibitem[{Hu et~al.(2026)Hu, Li, Yu, Liu, Ye, Chen and Huang}]{r1}
\bibinfo{author}{Hu, Z.}, \bibinfo{author}{Li, W.}, \bibinfo{author}{Yu, J.}, \bibinfo{author}{Liu, M.}, \bibinfo{author}{Ye, J.}, \bibinfo{author}{Chen, P.}, \bibinfo{author}{Huang, H.}, \bibinfo{year}{2026}.
\newblock \bibinfo{title}{3d building reconstruction from monocular remote sensing imagery via diffusion models and geometric priors}.
\newblock \bibinfo{journal}{ISPRS Journal of Photogrammetry and Remote Sensing} \bibinfo{volume}{232}, \bibinfo{pages}{124--137}.
\bibitem[{Huang et~al.(2024)Huang, Yu, Chen, Geiger and Gao}]{r4}
\bibinfo{author}{Huang, B.}, \bibinfo{author}{Yu, Z.}, \bibinfo{author}{Chen, A.}, \bibinfo{author}{Geiger, A.}, \bibinfo{author}{Gao, S.}, \bibinfo{year}{2024}.
\newblock \bibinfo{title}{2d gaussian splatting for geometrically accurate radiance fields}, in: \bibinfo{booktitle}{ACM SIGGRAPH 2024 conference papers}, pp. \bibinfo{pages}{1--11}.
\bibitem[{Huang et~al.(2026)Huang, Liu, Wan, Zheng, Zhang, Xiong, Pei and Zhang}]{r35}
\bibinfo{author}{Huang, X.}, \bibinfo{author}{Liu, X.}, \bibinfo{author}{Wan, Y.}, \bibinfo{author}{Zheng, Z.}, \bibinfo{author}{Zhang, B.}, \bibinfo{author}{Xiong, M.}, \bibinfo{author}{Pei, Y.}, \bibinfo{author}{Zhang, Y.}, \bibinfo{year}{2026}.
\newblock \bibinfo{title}{Skysplat: Generalizable 3d gaussian splatting from multi-temporal sparse satellite images}, in: \bibinfo{booktitle}{Proceedings of the AAAI Conference on Artificial Intelligence}, pp. \bibinfo{pages}{5158--5166}.
\bibitem[{Kerbl et~al.(2023)Kerbl, Kopanas, Leimk{\"u}hler, Drettakis et~al.}]{r3}
\bibinfo{author}{Kerbl, B.}, \bibinfo{author}{Kopanas, G.}, \bibinfo{author}{Leimk{\"u}hler, T.}, \bibinfo{author}{Drettakis, G.}, et~al., \bibinfo{year}{2023}.
\newblock \bibinfo{title}{3d gaussian splatting for real-time radiance field rendering.}
\newblock \bibinfo{journal}{ACM Trans. Graph.} \bibinfo{volume}{42}, \bibinfo{pages}{139--1}.
\bibitem[{Kuang and Liu(2026)}]{r21}
\bibinfo{author}{Kuang, H.}, \bibinfo{author}{Liu, J.}, \bibinfo{year}{2026}.
\newblock \bibinfo{title}{Gaussian entropy fields: Driving adaptive sparsity in 3d gaussian optimization}.
\newblock \bibinfo{journal}{ISPRS Journal of Photogrammetry and Remote Sensing} \bibinfo{volume}{236}, \bibinfo{pages}{273--285}.
\bibitem[{Lee et~al.(2025)Lee, Liu, Tsai, Chang, Wu, Chan, Zhao, Lin and Liu}]{r36}
\bibinfo{author}{Lee, J.Y.}, \bibinfo{author}{Liu, Y.R.}, \bibinfo{author}{Tsai, S.R.}, \bibinfo{author}{Chang, W.C.}, \bibinfo{author}{Wu, C.H.}, \bibinfo{author}{Chan, J.}, \bibinfo{author}{Zhao, Z.}, \bibinfo{author}{Lin, C.H.}, \bibinfo{author}{Liu, Y.L.}, \bibinfo{year}{2025}.
\newblock \bibinfo{title}{Skyfall-gs: Synthesizing immersive 3d urban scenes from satellite imagery}.
\newblock \bibinfo{journal}{arXiv preprint arXiv:2510.15869} .
\bibitem[{Li et~al.(2025)Li, Yao, Wu, Yue, Zhao, Qin, Garc{\'\i}a-Fern{\'a}ndez, Levers, Ralph and Zhu}]{r5}
\bibinfo{author}{Li, Z.}, \bibinfo{author}{Yao, S.}, \bibinfo{author}{Wu, T.}, \bibinfo{author}{Yue, Y.}, \bibinfo{author}{Zhao, W.}, \bibinfo{author}{Qin, R.}, \bibinfo{author}{Garc{\'\i}a-Fern{\'a}ndez, {\'A}.F.}, \bibinfo{author}{Levers, A.}, \bibinfo{author}{Ralph, J.}, \bibinfo{author}{Zhu, X.}, \bibinfo{year}{2025}.
\newblock \bibinfo{title}{Ulsr-gs: Urban large-scale surface reconstruction gaussian splatting with multi-view geometric consistency}.
\newblock \bibinfo{journal}{ISPRS Journal of Photogrammetry and Remote Sensing} \bibinfo{volume}{230}, \bibinfo{pages}{861--880}.
\bibitem[{Liu et~al.(2026)Liu, Niu, Qiu, Peng, Shang, Zhong and Ding}]{r14}
\bibinfo{author}{Liu, Z.}, \bibinfo{author}{Niu, S.}, \bibinfo{author}{Qiu, X.}, \bibinfo{author}{Peng, L.}, \bibinfo{author}{Shang, Y.}, \bibinfo{author}{Zhong, L.}, \bibinfo{author}{Ding, C.}, \bibinfo{year}{2026}.
\newblock \bibinfo{title}{A differentiable method for novel view sar image generation via 3d gaussian splatting}.
\newblock \bibinfo{journal}{ISPRS Journal of Photogrammetry and Remote Sensing} \bibinfo{volume}{231}, \bibinfo{pages}{167--195}.
\bibitem[{Lu et~al.(2024)Lu, Yu, Xu, Xiangli, Wang, Lin and Dai}]{r16}
\bibinfo{author}{Lu, T.}, \bibinfo{author}{Yu, M.}, \bibinfo{author}{Xu, L.}, \bibinfo{author}{Xiangli, Y.}, \bibinfo{author}{Wang, L.}, \bibinfo{author}{Lin, D.}, \bibinfo{author}{Dai, B.}, \bibinfo{year}{2024}.
\newblock \bibinfo{title}{Scaffold-gs: Structured 3d gaussians for view-adaptive rendering}, in: \bibinfo{booktitle}{Proceedings of the IEEE/CVF conference on computer vision and pattern recognition}, pp. \bibinfo{pages}{20654--20664}.
\bibitem[{Lu et~al.(2025)Lu, Chen, Yi, Chung, Wang and Hu}]{r41}
\bibinfo{author}{Lu, W.}, \bibinfo{author}{Chen, H.}, \bibinfo{author}{Yi, A.}, \bibinfo{author}{Chung, Y.Y.}, \bibinfo{author}{Wang, Z.}, \bibinfo{author}{Hu, K.}, \bibinfo{year}{2025}.
\newblock \bibinfo{title}{Learning fine-grained geometry for sparse-view splatting via cascade depth loss}.
\newblock \bibinfo{journal}{arXiv preprint arXiv:2505.22279} .
\bibitem[{Mar{\'\i} et~al.(2022)Mar{\'\i}, Facciolo and Ehret}]{r7}
\bibinfo{author}{Mar{\'\i}, R.}, \bibinfo{author}{Facciolo, G.}, \bibinfo{author}{Ehret, T.}, \bibinfo{year}{2022}.
\newblock \bibinfo{title}{Sat-nerf: Learning multi-view satellite photogrammetry with transient objects and shadow modeling using rpc cameras}, in: \bibinfo{booktitle}{Proceedings of the IEEE/CVF Conference on Computer Vision and Pattern Recognition}, pp. \bibinfo{pages}{1311--1321}.
\bibitem[{Mar{\'\i} et~al.(2023)Mar{\'\i}, Facciolo and Ehret}]{r30}
\bibinfo{author}{Mar{\'\i}, R.}, \bibinfo{author}{Facciolo, G.}, \bibinfo{author}{Ehret, T.}, \bibinfo{year}{2023}.
\newblock \bibinfo{title}{Multi-date earth observation nerf: The detail is in the shadows}, in: \bibinfo{booktitle}{Proceedings of the IEEE/CVF Conference on Computer Vision and Pattern Recognition}, pp. \bibinfo{pages}{2035--2045}.
\bibitem[{Mildenhall et~al.(2021)Mildenhall, Srinivasan, Tancik, Barron, Ramamoorthi and Ng}]{r2}
\bibinfo{author}{Mildenhall, B.}, \bibinfo{author}{Srinivasan, P.P.}, \bibinfo{author}{Tancik, M.}, \bibinfo{author}{Barron, J.T.}, \bibinfo{author}{Ramamoorthi, R.}, \bibinfo{author}{Ng, R.}, \bibinfo{year}{2021}.
\newblock \bibinfo{title}{Nerf: Representing scenes as neural radiance fields for view synthesis}.
\newblock \bibinfo{journal}{Communications of the ACM} \bibinfo{volume}{65}, \bibinfo{pages}{99--106}.
\bibitem[{M{\"u}ller et~al.(2022)M{\"u}ller, Evans, Schied and Keller}]{r12}
\bibinfo{author}{M{\"u}ller, T.}, \bibinfo{author}{Evans, A.}, \bibinfo{author}{Schied, C.}, \bibinfo{author}{Keller, A.}, \bibinfo{year}{2022}.
\newblock \bibinfo{title}{Instant neural graphics primitives with a multiresolution hash encoding}.
\newblock \bibinfo{journal}{ACM transactions on graphics (TOG)} \bibinfo{volume}{41}, \bibinfo{pages}{1--15}.
\bibitem[{Newcombe et~al.(2011)Newcombe, Izadi, Hilliges, Molyneaux, Kim, Davison, Kohi, Shotton, Hodges and Fitzgibbon}]{r46}
\bibinfo{author}{Newcombe, R.A.}, \bibinfo{author}{Izadi, S.}, \bibinfo{author}{Hilliges, O.}, \bibinfo{author}{Molyneaux, D.}, \bibinfo{author}{Kim, D.}, \bibinfo{author}{Davison, A.J.}, \bibinfo{author}{Kohi, P.}, \bibinfo{author}{Shotton, J.}, \bibinfo{author}{Hodges, S.}, \bibinfo{author}{Fitzgibbon, A.}, \bibinfo{year}{2011}.
\newblock \bibinfo{title}{Kinectfusion: Real-time dense surface mapping and tracking}, in: \bibinfo{booktitle}{2011 10th IEEE international symposium on mixed and augmented reality}, \bibinfo{organization}{Ieee}. pp. \bibinfo{pages}{127--136}.
\bibitem[{Ren et~al.(2024)Ren, Jiang, Lu, Yu, Xu, Ni and Dai}]{r17}
\bibinfo{author}{Ren, K.}, \bibinfo{author}{Jiang, L.}, \bibinfo{author}{Lu, T.}, \bibinfo{author}{Yu, M.}, \bibinfo{author}{Xu, L.}, \bibinfo{author}{Ni, Z.}, \bibinfo{author}{Dai, B.}, \bibinfo{year}{2024}.
\newblock \bibinfo{title}{Octree-gs: Towards consistent real-time rendering with lod-structured 3d gaussians}.
\newblock \bibinfo{journal}{arXiv preprint arXiv:2403.17898} .
\bibitem[{Shi et~al.(2025)Shi, Tang, Wu, Ji and Duan}]{r13}
\bibinfo{author}{Shi, C.}, \bibinfo{author}{Tang, F.}, \bibinfo{author}{Wu, Y.}, \bibinfo{author}{Ji, H.}, \bibinfo{author}{Duan, H.}, \bibinfo{year}{2025}.
\newblock \bibinfo{title}{Accurate and complete neural implicit surface reconstruction in street scenes using images and lidar point clouds}.
\newblock \bibinfo{journal}{ISPRS Journal of Photogrammetry and Remote Sensing} \bibinfo{volume}{220}, \bibinfo{pages}{295--306}.
\bibitem[{Tang et~al.(2024)Tang, Ye, Ye, Lin, Zhou, Chen and Ouyang}]{r27}
\bibinfo{author}{Tang, S.}, \bibinfo{author}{Ye, W.}, \bibinfo{author}{Ye, P.}, \bibinfo{author}{Lin, W.}, \bibinfo{author}{Zhou, Y.}, \bibinfo{author}{Chen, T.}, \bibinfo{author}{Ouyang, W.}, \bibinfo{year}{2024}.
\newblock \bibinfo{title}{Hisplat: Hierarchical 3d gaussian splatting for generalizable sparse-view reconstruction}.
\newblock \bibinfo{journal}{arXiv preprint arXiv:2410.06245} .
\bibitem[{Wang et~al.(2004)Wang, Bovik, Sheikh and Simoncelli}]{r42}
\bibinfo{author}{Wang, Z.}, \bibinfo{author}{Bovik, A.C.}, \bibinfo{author}{Sheikh, H.R.}, \bibinfo{author}{Simoncelli, E.P.}, \bibinfo{year}{2004}.
\newblock \bibinfo{title}{Image quality assessment: from error visibility to structural similarity}.
\newblock \bibinfo{journal}{IEEE transactions on image processing} \bibinfo{volume}{13}, \bibinfo{pages}{600--612}.
\bibitem[{Xiang et~al.(2026)Xiang, Zhang, Li, Yang, Gao, Liu, Zhao, Li and Huang}]{r22}
\bibinfo{author}{Xiang, H.}, \bibinfo{author}{Zhang, F.}, \bibinfo{author}{Li, X.}, \bibinfo{author}{Yang, C.}, \bibinfo{author}{Gao, Y.}, \bibinfo{author}{Liu, W.}, \bibinfo{author}{Zhao, L.}, \bibinfo{author}{Li, D.}, \bibinfo{author}{Huang, X.}, \bibinfo{year}{2026}.
\newblock \bibinfo{title}{Gaussiancraft: Fine-grained 3d gaussians for efficient large-scene surface reconstruction}.
\newblock \bibinfo{journal}{ISPRS Journal of Photogrammetry and Remote Sensing} \bibinfo{volume}{235}, \bibinfo{pages}{651--667}.
\bibitem[{Xu et~al.(2025a)Xu, Peng, Wang, Blum, Barath, Geiger and Pollefeys}]{r26}
\bibinfo{author}{Xu, H.}, \bibinfo{author}{Peng, S.}, \bibinfo{author}{Wang, F.}, \bibinfo{author}{Blum, H.}, \bibinfo{author}{Barath, D.}, \bibinfo{author}{Geiger, A.}, \bibinfo{author}{Pollefeys, M.}, \bibinfo{year}{2025}a.
\newblock \bibinfo{title}{Depthsplat: Connecting gaussian splatting and depth}, in: \bibinfo{booktitle}{Proceedings of the Computer Vision and Pattern Recognition Conference}, pp. \bibinfo{pages}{16453--16463}.
\bibitem[{Xu et~al.(2025b)Xu, Gao and Shan}]{r29}
\bibinfo{author}{Xu, J.}, \bibinfo{author}{Gao, S.}, \bibinfo{author}{Shan, Y.}, \bibinfo{year}{2025}b.
\newblock \bibinfo{title}{Freesplatter: Pose-free gaussian splatting for sparse-view 3d reconstruction}, in: \bibinfo{booktitle}{Proceedings of the IEEE/CVF International Conference on Computer Vision}, pp. \bibinfo{pages}{25442--25452}.
\bibitem[{Yao et~al.(2018)Yao, Luo, Li, Fang and Quan}]{r39}
\bibinfo{author}{Yao, Y.}, \bibinfo{author}{Luo, Z.}, \bibinfo{author}{Li, S.}, \bibinfo{author}{Fang, T.}, \bibinfo{author}{Quan, L.}, \bibinfo{year}{2018}.
\newblock \bibinfo{title}{Mvsnet: Depth inference for unstructured multi-view stereo}, in: \bibinfo{booktitle}{Proceedings of the European conference on computer vision (ECCV)}, pp. \bibinfo{pages}{767--783}.
\bibitem[{Yao et~al.(2026)Yao, Zhang, Zhang, Gao, Peng, Li, Wang and Wang}]{r15}
\bibinfo{author}{Yao, Y.}, \bibinfo{author}{Zhang, B.}, \bibinfo{author}{Zhang, W.}, \bibinfo{author}{Gao, L.}, \bibinfo{author}{Peng, D.}, \bibinfo{author}{Li, B.}, \bibinfo{author}{Wang, Y.}, \bibinfo{author}{Wang, B.}, \bibinfo{year}{2026}.
\newblock \bibinfo{title}{Arsgaussian: 3d gaussian splatting with lidar for aerial remote sensing novel view synthesis}.
\newblock \bibinfo{journal}{ISPRS Journal of Photogrammetry and Remote Sensing} \bibinfo{volume}{231}, \bibinfo{pages}{288--306}.
\bibitem[{Ye et~al.(2024)Ye, Li, Liu, Qiao and Dou}]{r18}
\bibinfo{author}{Ye, Z.}, \bibinfo{author}{Li, W.}, \bibinfo{author}{Liu, S.}, \bibinfo{author}{Qiao, P.}, \bibinfo{author}{Dou, Y.}, \bibinfo{year}{2024}.
\newblock \bibinfo{title}{Absgs: Recovering fine details in 3d gaussian splatting}, in: \bibinfo{booktitle}{Proceedings of the 32nd ACM international conference on multimedia}, pp. \bibinfo{pages}{1053--1061}.
\bibitem[{Yu et~al.(2024)Yu, Sattler and Geiger}]{r23}
\bibinfo{author}{Yu, Z.}, \bibinfo{author}{Sattler, T.}, \bibinfo{author}{Geiger, A.}, \bibinfo{year}{2024}.
\newblock \bibinfo{title}{Gaussian opacity fields: Efficient adaptive surface reconstruction in unbounded scenes}.
\newblock \bibinfo{journal}{ACM Transactions on Graphics (ToG)} \bibinfo{volume}{43}, \bibinfo{pages}{1--13}.
\bibitem[{Zhang et~al.(2025a)Zhang, Zou, Li, Yi and Wang}]{r25}
\bibinfo{author}{Zhang, C.}, \bibinfo{author}{Zou, Y.}, \bibinfo{author}{Li, Z.}, \bibinfo{author}{Yi, M.}, \bibinfo{author}{Wang, H.}, \bibinfo{year}{2025}a.
\newblock \bibinfo{title}{Transplat: Generalizable 3d gaussian splatting from sparse multi-view images with transformers}, in: \bibinfo{booktitle}{Proceedings of the AAAI Conference on Artificial Intelligence}, pp. \bibinfo{pages}{9869--9877}.
\bibitem[{Zhang et~al.(2019)Zhang, Snavely and Sun}]{r40}
\bibinfo{author}{Zhang, K.}, \bibinfo{author}{Snavely, N.}, \bibinfo{author}{Sun, J.}, \bibinfo{year}{2019}.
\newblock \bibinfo{title}{Leveraging vision reconstruction pipelines for satellite imagery}, in: \bibinfo{booktitle}{Proceedings of the IEEE/CVF International Conference on Computer Vision Workshops}, pp. \bibinfo{pages}{0--0}.
\bibitem[{Zhang and Rupnik(2023)}]{r31}
\bibinfo{author}{Zhang, L.}, \bibinfo{author}{Rupnik, E.}, \bibinfo{year}{2023}.
\newblock \bibinfo{title}{Sparsesat-nerf: Dense depth supervised neural radiance fields for sparse satellite images}.
\newblock \bibinfo{journal}{arXiv preprint arXiv:2309.00277} .
\bibitem[{Zhang et~al.(2025b)Zhang, Wysocki and Jutzi}]{r6}
\bibinfo{author}{Zhang, Q.}, \bibinfo{author}{Wysocki, O.}, \bibinfo{author}{Jutzi, B.}, \bibinfo{year}{2025}b.
\newblock \bibinfo{title}{Gs4buildings: Prior-guided gaussian splatting for 3d building reconstruction}.
\newblock \bibinfo{journal}{arXiv preprint arXiv:2508.07355} .
\bibitem[{Zhang et~al.(2018)Zhang, Isola, Efros, Shechtman and Wang}]{r43}
\bibinfo{author}{Zhang, R.}, \bibinfo{author}{Isola, P.}, \bibinfo{author}{Efros, A.A.}, \bibinfo{author}{Shechtman, E.}, \bibinfo{author}{Wang, O.}, \bibinfo{year}{2018}.
\newblock \bibinfo{title}{The unreasonable effectiveness of deep features as a perceptual metric}, in: \bibinfo{booktitle}{Proceedings of the IEEE conference on computer vision and pattern recognition}, pp. \bibinfo{pages}{586--595}.
\bibitem[{Zhang et~al.(2024a)Zhang, Li and Wei}]{r33}
\bibinfo{author}{Zhang, T.}, \bibinfo{author}{Li, Y.}, \bibinfo{author}{Wei, X.}, \bibinfo{year}{2024}a.
\newblock \bibinfo{title}{rpcprf: Generalizable mpi neural radiance field for satellite camera with single and sparse views}.
\newblock \bibinfo{journal}{IEEE Transactions on Geoscience and Remote Sensing} \bibinfo{volume}{62}, \bibinfo{pages}{1--15}.
\bibitem[{Zhang et~al.(2024b)Zhang, Zhou, Li and Wei}]{r32}
\bibinfo{author}{Zhang, T.}, \bibinfo{author}{Zhou, Y.}, \bibinfo{author}{Li, Y.}, \bibinfo{author}{Wei, X.}, \bibinfo{year}{2024}b.
\newblock \bibinfo{title}{Satensorf: Fast satellite tensorial radiance field for multidate satellite imagery of large size}.
\newblock \bibinfo{journal}{IEEE Transactions on Geoscience and Remote Sensing} \bibinfo{volume}{62}, \bibinfo{pages}{1--15}.

\end{thebibliography}



\end{document}